\documentclass[10pt,journal,compsoc]{IEEEtran}
\usepackage{times}
\usepackage{epsfig}
\usepackage{graphicx}
\usepackage{amsmath}
\usepackage{amssymb}
\usepackage{hyperref}
\usepackage{enumerate}
\usepackage{enumitem}
\usepackage{color}
\usepackage{comment}
\usepackage{url}
\usepackage{xcolor}
\usepackage{tabu}
\usepackage{booktabs}
\usepackage{makecell}
\usepackage{wrapfig}
\usepackage{breakcites}

\usepackage{algorithm}
\usepackage[noend]{algpseudocode}

\makeatletter
\def\BState{\State\hskip-\ALG@thistlm}
\makeatother

\ifCLASSOPTIONcompsoc
  \usepackage[nocompress]{cite}
\else
  \usepackage{cite}
\fi
\ifCLASSINFOpdf
\else
\fi
\begin{document}
%
\title{Deep Back-Projection Networks for Single Image Super-resolution}
%
%
%
%

\author{Muhammad~Haris,
        Greg~Shakhnarovich,
        and~Norimichi~Ukita,~\IEEEmembership{Member,~IEEE}
\IEEEcompsocitemizethanks{\IEEEcompsocthanksitem M. Haris and N. Ukita are with Intelligent Information Media Lab, Toyota Technological Institute (TTI), Nagoya,
Japan, 468-8511.\protect\\
E-mail: muhammad.haris@bukalapak.com* and ukita@toyota-ti.ac.jp (*he is currently working at Bukalapak in Indonesia)
\IEEEcompsocthanksitem G. Shakhnarovich is with TTI at Chicago, US.
E-mail: greg@ttic.edu}
\thanks{Manuscript received -; revised -.}}

\IEEEtitleabstractindextext{%
\begin{abstract}
Previous feed-forward architectures of recently proposed deep
super-resolution networks learn the features of low-resolution
inputs and the non-linear mapping from those to a high-resolution
output. However, this approach does not fully address the mutual
dependencies of low- and high-resolution images. 
We propose Deep Back-Projection Networks (DBPN), 
the winner of two image super-resolution challenges (NTIRE2018 and PIRM2018),
that exploit iterative up- and down-sampling layers.
These layers are formed as a unit providing an error feedback mechanism for projection errors.
We construct mutually-connected up- and down-sampling
units each of which represents different types of low- and high-resolution components. 
We also show that extending this idea to demonstrate a new insight towards more efficient network design substantially, such as parameter sharing on the projection module and transition layer on projection step.
The experimental results yield superior results and in particular
establishing new state-of-the-art results across multiple data sets, especially for large scaling factors
such as $8\times$.
\end{abstract}

}

\maketitle

\IEEEdisplaynontitleabstractindextext

%
\IEEEpeerreviewmaketitle

\IEEEraisesectionheading{\section{Introduction}
\label{sec:introduction}}

\IEEEPARstart{S}{ignificant} progress in deep neural network (DNN) for
vision~\cite{huang2017densely,he2015deep,denton2015deep,shrivastava2016learning,larsson2016fractalnet,radford2015unsupervised,IMKDB17}
has recently been propagating to the field of super-resolution (SR)~\cite{johnson2016perceptual,tao2017detail,sajjadi2018frame,Haris17,Kim_2016_VDSR,LapSRN,Tai-DRRN-2017,zhang2018image,park2018srfeat}. 

Single image SR (SISR) is an ill-posed inverse problem where the aim is to
recover a high-resolution (HR) image from a low-resolution (LR)
image. A currently typical approach is to construct an HR image by
learning non-linear LR-to-HR mapping, implemented as a DNN~\cite{dong2016image, dong2016accelerating, shi2016real, LapSRN, Kim_2016_VDSR, kim2016deeply, Tai-DRRN-2017} and non DNN~\cite{schulter2015fast,haris2017first,timofte2014a+,Haris17Sparse}. 
For DNN approach, the networks compute a sequence of feature maps from the LR image, culminating with
one or more upsampling layers to increase resolution and finally
construct the HR image. In contrast to this purely feed-forward
approach, the human visual system is believed to use a feedback connection
to simply guide the task for the relevant
results~\cite{felleman1991distributed, kravitz2013ventral,
  lamme2000distinct}. Perhaps hampered by lack of such feedback,
the current SR networks with only feed-forward connections have difficulty in representing the LR-to-HR relation, especially for large scaling factors.

On the other hand, feedback connections were used effectively by one
of the early SR algorithms, the iterative
back-projection~\cite{irani93}. It iteratively computes the
reconstruction error, then uses it to refine the HR image. 
Although it has been proven to improve the image quality,
results still suffers from ringing and chessboard
artifacts~\cite{dai2007bilateral}. Moreover, this method is sensitive to
choices of parameters such as the number of iterations and the blur
operator, leading to variability in results.



\begin{figure}[t!]
\centering
\includegraphics[width=8cm]{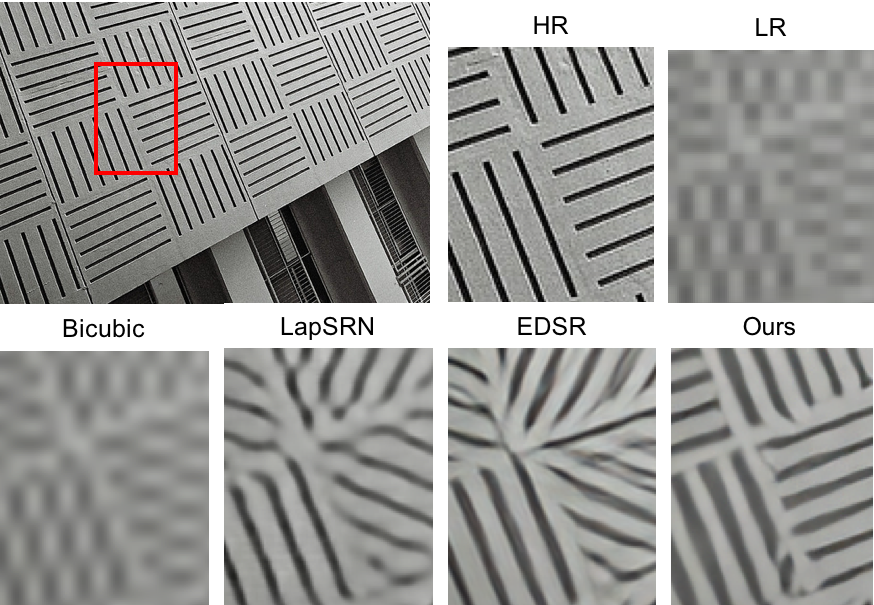}
\caption{Super-resolution result on $8\times$ enlargement. PSNR: LapSRN~\cite{LapSRN} (15.25 dB), EDSR~\cite{Lim_2017_CVPR_Workshops} (15.33 dB), and Ours~\cite{haris2018deep} (16.63 dB).}
\label{figure:intro}
\end{figure}

\begin{figure*}[t!]
\centering
\includegraphics[width=15cm]{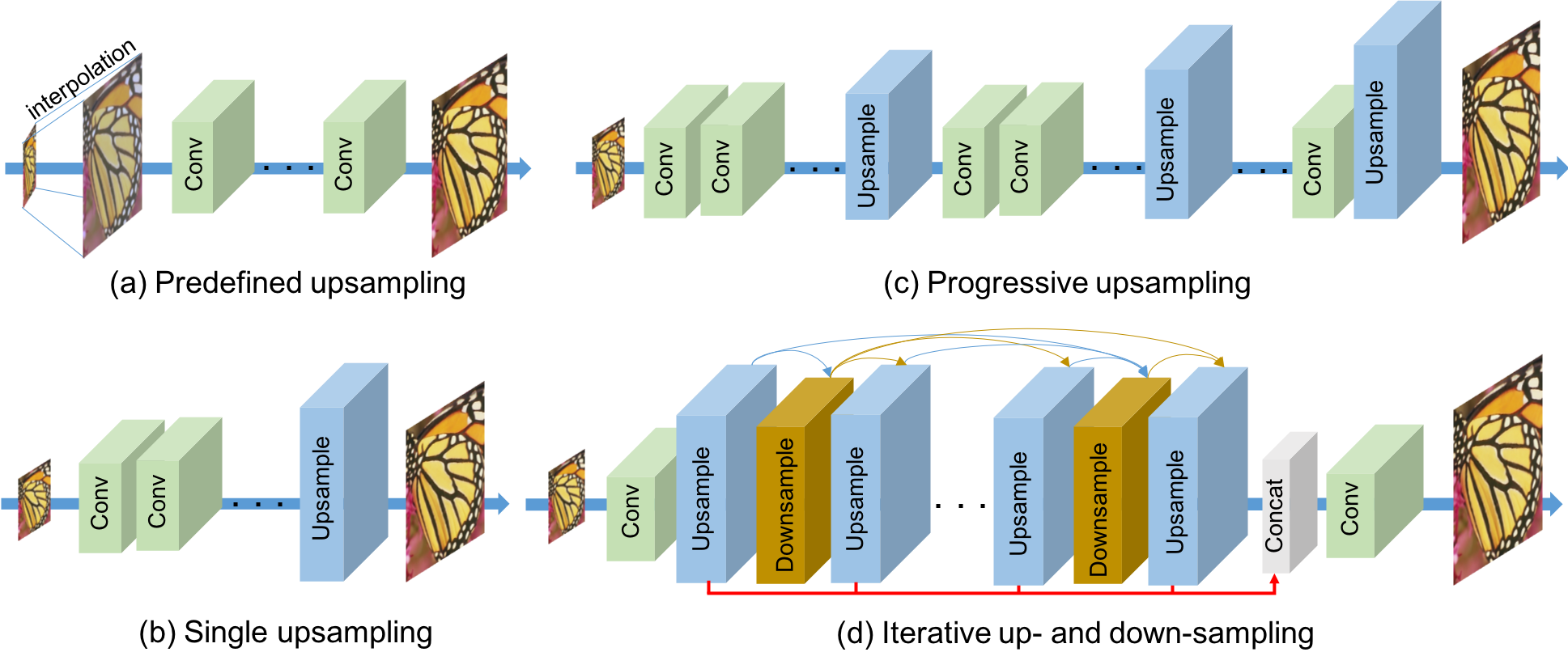}\vspace{-1em}
\caption{Comparisons of Deep Network SR. (a) Predefined upsampling (e.g., SRCNN~\cite{dong2016image}, VDSR~\cite{Kim_2016_VDSR}, DRRN~\cite{Tai-DRRN-2017}) commonly uses the conventional interpolation, such as Bicubic, to upscale LR input images before entering the network. (b) Single upsampling (e.g., FSRCNN~\cite{dong2016accelerating}, ESPCN~\cite{shi2016real}) propagates the LR features, then construct the SR image at the last step. (c) Progressive upsampling uses a Laplacian pyramid network to gradually predict SR images~\cite{LapSRN}. (d) Iterative up- and down-sampling approach is proposed by our DBPN that exploit the mutually connected up- (blue box) and down-sampling (gold box) units to obtain numerous HR feature maps in different depths.}
\label{figure:multi_upsampling}
\end{figure*}

Inspired by~\cite{irani93}, we construct an end-to-end
trainable architecture based on the idea of iterative up- and
down-sampling layers: Deep Back-Projection Networks (DBPN). Our networks are not only able to remove the ringing and chessboard effect but also successfully perform large scaling factors, as shown
in~Fig.~\ref{figure:intro}. 
Furthermore, DBPN has been proven by winning SISR challenges. On NTIRE2018~\cite{timofte2018ntire}, DBPN is the $1^{st}$ winner on track 8$\times$ Bicubic downscaling. On PIRM2018~\cite{pirm2018}, DBPN got $1^{st}$ on Region 2, $3^{rd}$ on Region 1, and $5^{th}$ on Region 3.


Our work provides the following contributions:

\noindent(1) \textbf{Iterative up- and down-sampling units}. 
Most of the existing methods extract the feature maps with the same size of LR input image, then these feature maps are finally upsampled to produce SR image directly or progressively.
 However, DBPN performed iterative up- and down- sampling layers so that it can capture features at each resolution that are helpful in recovering features at other resolution in a single framework. 
Our networks focus not only on generating variants of the HR features using the up-sampling unit but also on projecting it back to the LR spaces using the down-sampling unit. 
It is shown in~Fig.~\ref{figure:multi_upsampling} (d), alternating between up- (blue box) and down-sampling (gold box) units, which represent the mutual relation of LR and HR features. 
Each set of up- and down-sampling layers achieve data augmentation in the feature space to represent a certain set of features at LR and HR resolution. So, multiple sets of up- and down-sampling layers represent multiple feature maps that are useful for producing the SR image of an input LR image. That is, even if all features extracted by these layers are produced from the single input LR image, these features can be different from each other by training DBPN so that their variety helps to improve its output (i.e., SR image). The detailed explanation can be seen in Section~\ref{subsec:network architecture}.

\noindent(2) \textbf{Error feedback}. We propose an iterative
error-correcting feedback mechanism for SR, which calculates both up- and down-projection errors to guide the reconstruction for obtaining better results. 
Here, the projection errors are used to refine the initial features in early layers. 
The detailed explanation can be seen in Section~\ref{subsec:projection_unit}.

\noindent(3) \textbf{Deep concatenation}. Our networks represent different types of LR and HR components produced by each up- and down-sampling unit. 
This ability enables the networks to reconstruct the HR image using concatenation of the HR feature maps from all of the up-sampling units. 
Our reconstruction can directly use different types of HR feature maps from different depths without propagating them through the other layers as shown by the red arrows in~Fig.~\ref{figure:multi_upsampling} (d).


In this work, we make the following extensions to demonstrate a new insight towards more efficient network design substantially compare to our early results~\cite{haris2018deep}. 
This manuscript focuses on optimized DBPN architecture with advanced design methodologies. 
The detailed explanation can be seen in Section~\ref{subsec:recurrent_dbpn}. 
Based on this experiment, we found several technical contributions as follows.

\noindent(1) \textbf{Parameter sharing on projection module}. 
Due to a large amount of parameters, it is hard to train deeper DBPN. 
Based on our experiments, the deepest setting uses $T=10$. 
However, using parameter sharing, we can avoid the increasing number of parameters, while widening the receptive field. 
The effectiveness of parameter sharing was shown in Table~\ref{tab:psnr_variant} where DBPN-R64-10 has better performance than D-DBPN while reducing the number of parameters by 10$\times$.

\noindent(2) \textbf{Transition layer on projection step}.
Inspired by dense connection network~\cite{huang2017densely}, we use transition layer to create multiple projection step. 
The last up- and down-projection units were used as transition layer. 
The last up-projection was used to produce final HR feature-maps for each iteration, and the last down-projection was used to produce the next input for the next iteration.
Table~\ref{tab:psnr_variant} shows that DBPN-RES-MR64-3 outperforms previous setting, D-DBPN, by a large margin. DBPN-RES-MR64-3 successfully improves D-DBPN performance by 0.7 dB on Urban100 without increasing the model parameter.


\section{Related Work}
\label{sec:related}
\subsection{Image super-resolution using deep networks}
Deep Networks SR can be primarily divided into four types as shown in~Fig.~\ref{figure:multi_upsampling}.

(a) \textbf{Predefined upsampling} commonly uses interpolation as the upsampling operator to produce a middle resolution (MR) image. This scheme was proposed by SRCNN~\cite{dong2016image} to learn MR-to-HR non-linear mapping with simple convolutional layers as in non deep learning based approaches~\cite{timofte2014a+,schulter2015fast}. Later, the improved networks exploited residual learning~\cite{Kim_2016_VDSR,Tai-DRRN-2017} and recursive layers~\cite{kim2016deeply}. However, this approach has higher computation because the input is the MR image which has the same size as the HR image.

(b) \textbf{Single upsampling} offers a simple way to increase the resolution. This approach was firstly proposed by FSRCNN~\cite{dong2016accelerating} and ESPCN~\cite{shi2016real}. 
These methods have been proven effective to increase the resolution and replace predefined operators. 
Further improvements include residual network~\cite{Lim_2017_CVPR_Workshops}, dense connection~\cite{zhang2018residual}, and channel attention~\cite{zhang2018image}
However, they fail to learn complicated mapping of LR-to-HR image, especially on large scaling factors, due to limited feature maps from the LR image.
This problem opens the opportunities to propose the mutual relation from LR-to-HR image that can preserve HR components better.

(c) \textbf{Progressive upsampling} was recently proposed in
LapSRN~\cite{LapSRN}. It progressively reconstructs the multiple SR
images with different scales in one feed-forward network. For the sake
of simplification, we can say that this network is a stacked of single upsampling networks which only relies on limited LR feature maps. 
Due to this fact, LapSRN is outperformed even by our shallow networks especially for large scaling factors such as $8\times$ in experimental results.

(d) \textbf{Iterative up- and down-sampling} is proposed by our networks~\cite{haris2018deep}. We focus on increasing the sampling rate of HR feature maps in different depths from iterative up- and down-sampling layers, then, distribute the tasks to calculate the reconstruction error on each unit. This scheme enables the networks to preserve the HR components by learning various up- and down-sampling operators while generating deeper features.

\subsection{Feedback networks} 
Rather than learning a non-linear mapping of input-to-target space in one step, the feedback networks compose the prediction process into multiple steps which allow the model to have a self-correcting procedure. Feedback procedure has been implemented in various computing tasks~\cite{carreira2016human,ross2011learning,tu2010auto,li2016iterative,zamir2016feedback, shrivastava2016contextual,lotter2016deep}.

In the context of human pose estimation, Carreira et al.~\cite{carreira2016human} proposed an iterative error feedback by
iteratively estimating and applying a correction to the current
estimation. PredNet~\cite{lotter2016deep} is an unsupervised recurrent
network to predictively code the future frames by recursively feeding
the predictions back into the model. For image segmentation, Li et
al.~\cite{li2016iterative} learn implicit shape priors and use them to
improve the prediction. However, to our knowledge, feedback procedures
have not been implemented to SR.

\subsection{Adversarial training}
Adversarial training, such as with Generative Adversarial Networks (GANs)~\cite{goodfellow2014generative} has been applied to various image reconstruction problems~\cite{ledig2016photo, sajjadi2016enhancenet, radford2015unsupervised, denton2015deep, johnson2016perceptual}. For the SR task, Johnson et al.~\cite{johnson2016perceptual} introduced perceptual losses based on high-level features extracted from pre-trained networks. Ledig et al.~\cite{ledig2016photo} proposed SRGAN which is considered as a single upsampling method. It proposed the natural image manifold that is able to create photo-realistic images by specifically formulating a loss function based on the euclidian distance between feature maps extracted from VGG19~\cite{simonyan2014very}. Our networks can be extended with the adversarial training. The detailed explanation is available in Section~\ref{sec:perceptual}.

\subsection{Back-projection} 
Back-projection~\cite{irani93} is an efficient iterative
procedure to minimize the reconstruction error. Previous studies have
proven the effectiveness of back-projection~\cite{zhao2017iterative,
  haris2017first, dong2009nonlocal, timofte2016seven}. Originally,
back-projection in SR was designed for the case with multiple LR inputs. However, given only one LR input image, the reconstruction procedure can be obtained by upsampling the LR image using multiple upsampling operators and calculate the reconstruction error iteratively~\cite{dai2007bilateral}. Timofte et al.~\cite{timofte2016seven} mentioned that back-projection could improve the quality of the SR images. Zhao et al.~\cite{zhao2017iterative} proposed a method to refine high-frequency texture details with an iterative projection process. However, the initialization which leads to an optimal solution remains unknown. Most of the previous studies involve constant and unlearned predefined parameters such as blur operator and number of iteration. 



\section{Deep Back-Projection Networks} 
\label{sec:proposed}
Let $I^h$ and $I^l$ be HR and LR image with  $(M^h \times N^h)$ and $(
M^l \times N^l)$, respectively, where $M^l < M^h$ and $N^l <
N^h$. The main building block of our proposed DBPN architecture is the
projection unit, which is trained (as part of the end-to-end training
of the SR system) to map either an LR feature map to an HR map
(up-projection), or an HR map to an LR map (down-projection). 

\subsection{Iterative back-projection}
\label{subsec:back_projection}
Back-projection is originally designed with multiple LR inputs~\cite{irani93}. Given only one LR input image, the iterative back-projection~\cite{dai2007bilateral} can be summarized as follows.

\begin{align}\label{eq:back-projection}
&\text{scale up:}&\hat{I}^h_t &= (I^l_t * p)\uparrow_{s},\\
&\text{scale down:}&\hat{I}^l_t &= (\hat{I}^h_t * g) \downarrow_{s},\\
&\text{residual:}&e^l_t &= I^l_t - \hat{I}^l_t,\\
&\text{scale residual up:}&e^h_t &= (e^l_t* p)\uparrow_{s},\\
&\text{output image:}&\hat{I}^h_{t+1} &= \hat{I}^h_t + e^h_t&
\end{align}

where $\hat{I}^h_{t+1}$ is the final SR image at the $t$-th iteration, 
$p$ is a constant back-projection kernel and $g$ is a single blur filter, 
$\uparrow_{s}$ and $\downarrow_{s}$ are the up- and down-sampling operator, respectively.

The traditional back-projection relies on constant and unlearned predefined parameters such as single sampling filter and blur operator. 
To extend this algorithm, our proposal preserves the HR components by learned various up- and down-sampling operators and generates deeper features to construct numerous pair of LR-and-HR feature maps. 
We develop an end-to-end trainable architecture which focuses to guide the SR task using mutually connected up- and down-sampling layers to learn non-linear mutual relation of LR-to-HR components. 
The mutual relation between LR and HR components is constructed by creating iterative up- and down-projection layers where the up-projection unit generates HR feature maps, then the down-projection unit projects it back to the LR spaces as shown in~Fig.~\ref{figure:multi_upsampling} (d).

\subsection{Projection units}
\label{subsec:projection_unit}

The up-projection unit is defined as follows:
\begin{align}\label{eq:up-projection}
&\text{scale up:}&H^t_0 &= (L^{t-1} * p_{t})\uparrow_{s},\\
&\text{scale down:}&L^t_0 &= (H^t_0 * g_{t}) \downarrow_{s},\\
&\text{residual:}&e^l_t &= L^t_0 - L^{t-1},\\
&\text{scale residual up:}&H^t_1 &= (e^l_t* q_{t})\uparrow_{s},\\
&\text{output feature map:}&H^t &= H^t_0 + H^t_1&
\end{align}
where * is the spatial convolution operator, $\uparrow_{s}$ and $\downarrow_{s}$ are, respectively, the up- and
down-sampling operator with scaling factor $s$, and $p_t,
g_t, q_t$ are (de)convolutional layers at stage $t$.

\begin{figure}[t!]
\centering
\includegraphics[width=8cm]{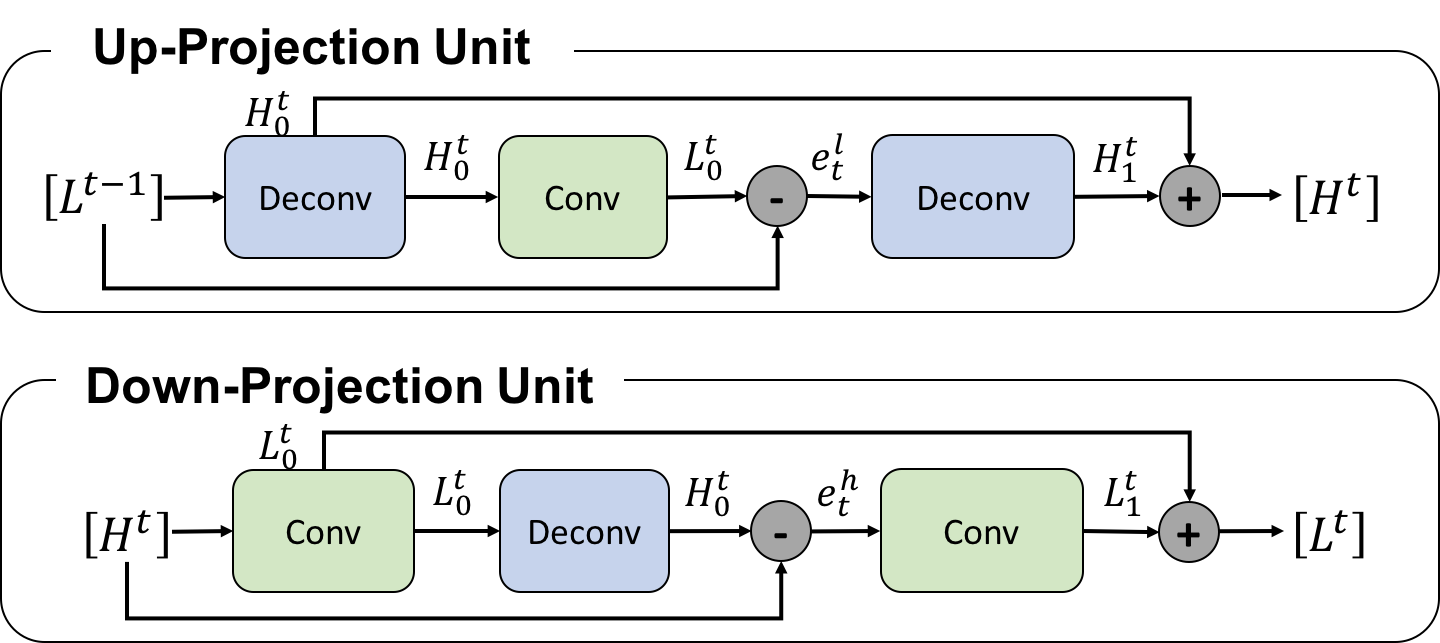}\vspace{-1em}
\caption{Proposed up- and down-projection units in the DBPN. These units produce residual $e$ between the initial features and the reconstructed features, then fuse it back by summing it to the initial features.}
\label{figure:projection_unit}
\end{figure}

The up-projection unit, illustrated in the upper part of~Fig.~\ref{figure:projection_unit}, takes the previously computed LR feature map
$L^{t-1}$ as input, and maps it to an (intermediate) HR map $H^t_0$;
then it attempts to map it back to LR map $L^t_0$
(``back-project''). The residual (difference) $e^l_t$ between the observed LR map
$L^{t-1}$ and the reconstructed $L^t_0$ is mapped to HR again,
producing a new intermediate (residual) map $H^t_1$; the final output of the unit,
the HR map $H^t$, is obtained by summing the two intermediate HR maps.

The down-projection unit, illustrated in the lower part of~Fig.~\ref{figure:projection_unit}, is defined very similarly, but now its job is
to map its input HR map $H^t$ to the LR map $L^t$.
\begin{align}\label{eq:down-projection}
&\text{scale down:}&L^t_0 &= (H^{t} * g'_{t})\downarrow_{s},\\
&\text{scale up:}&H^t_0 &= (L^t_0 * p'_{t}) \uparrow_{s},\\
&\text{residual:}&e^h_t &= H^t_0 - H^{t},\\
&\text{scale residual down:}&L^t_1 &= (e^h_t * g'_{t})\downarrow_{s},\\
&\text{output feature map:}&L^t &= L^t_0 + L^t_1\label{eq:proj-last}
\end{align}

We organize projection units in a series of \emph{stages}, alternating between $H$ and $L$.
These projection units can be understood as a self-correcting procedure which feeds a projection error to the sampling layer and iteratively changes the solution by feeding back the projection error.

The projection unit uses large sized filters such as $8\times 8$ and $12\times 12$. In the previous approaches, the use of large-sized filters is avoided because it can slow down the convergence speed and might produce sub-optimal results. 
However, the iterative up- and down-sampling units enable the mutual relation between LR and HR components.
These units also take benefit of large receptive fields to perform better performance especially on large scaling factor where the significant amount of pixels is needed.

\subsection{Network architecture}
\label{subsec:network architecture}
\begin{figure*}[t!]
\centering
\includegraphics[width=14cm]{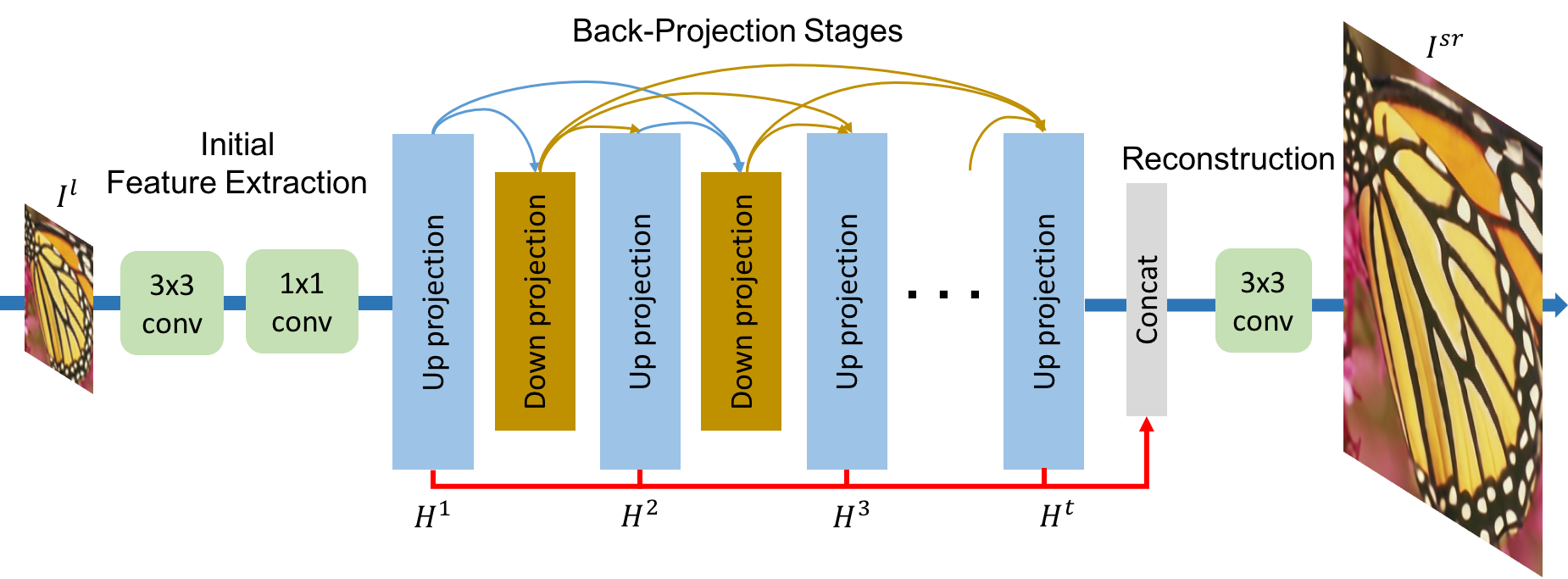}\vspace{-1.5em}
\caption{An implementation of DBPN for super-resolution which exploits densely connected projection unit to encourage feature reuse.}
\label{figure:proposed_network}
\end{figure*}

The proposed DBPN is illustrated in~Fig.~\ref{figure:proposed_network}. It can be divided into three
parts: initial feature extraction, projection, and reconstruction, as
described below. Here, let \texttt{conv}$(f,n)$ be a convolutional layer,
where $f$ is the filter size and $n$ is the number of filters. 

\begin{enumerate}
\item \textbf{Initial feature extraction}. We construct initial LR
  feature-maps $L^0~\in~\mathbb{R}^{M^l \times N^l \times n_{0}}$ from the input using \texttt{conv}$(3,n_{0})$. 
  Then \texttt{conv}$(1,n_R )$ is used to reduce the dimension from $n_{0}$ to $n_R$ before
  entering projection step where $n_0$ is the number of filters used in the initial LR features extraction and $n_R$ is the number of filters used in each projection unit.
\item \textbf{Back-projection stages}. Following initial feature
  extraction is a sequence of projection units, alternating between
  construction of LR and HR feature maps ($L^t~\in~\mathbb{R}^{M^l \times N^l \times n_R}$ and $H^t~\in~\mathbb{R}^{M^h \times N^h \times n_R}$).
  Later, it further improves by dense connection where each unit has access to the outputs of all previous units (Section~\ref{subsec:dense_projection_unit}).

\item \textbf{Reconstruction}. Finally, the target HR image is
  reconstructed as $I^{sr}=f_{Rec}([H^{1},H^{2},
  ..., H^{t}]),$ where $f_{Rec}$ use \texttt{conv}$(3,3)$ as
  reconstruction and $[H^{1},H^{2},
  ..., H^{t}]$ refers to the concatenation of the feature-maps
  produced in each up-projection unit which called as deep concatenation.
\end{enumerate}

Due to the definitions of these building blocks, our network
architecture is modular. We can easily define and train networks with
different numbers of stages, controlling the depth. For a network with
$T$ stages, we have the initial extraction stage (2 layers), and then
$T$ up-projection units and $T-1$ down-projection units, each with 3
layers, followed by the reconstruction (one more layer). 
However, for the dense projection unit, we add \texttt{conv}$(1,n_R)$ in each projection unit, except the first three units as mentioned in Section~\ref{subsec:dense_projection_unit}.
\section{The Variants of DBPN}
\label{sec:variant}
In this section, we show how DBPN can be modified to apply the latest deep learning trends. 

\subsection{Dense projection units}
\label{subsec:dense_projection_unit}
The dense inter-layer connectivity pattern in DenseNets~\cite{huang2017densely} has been shown to 
alleviate the vanishing-gradient problem, produce improved features,
and encourage feature reuse. Inspired by this we propose to improve
DBPN, by introducing dense connections in the projection units called,
yielding Dense DBPN.

Unlike the original DenseNets, we avoid dropout and batch norm, which are not suitable for SR, because they remove the range flexibility of the features~\cite{Lim_2017_CVPR_Workshops}.
Instead, we use $1 \times 1$ convolution layer as the bottleneck layer for feature pooling and dimensional reduction~\cite{szegedy2015going,Haris17} before entering the projection unit.

\begin{figure}[t!]
\centering
\includegraphics[width=8.5cm]{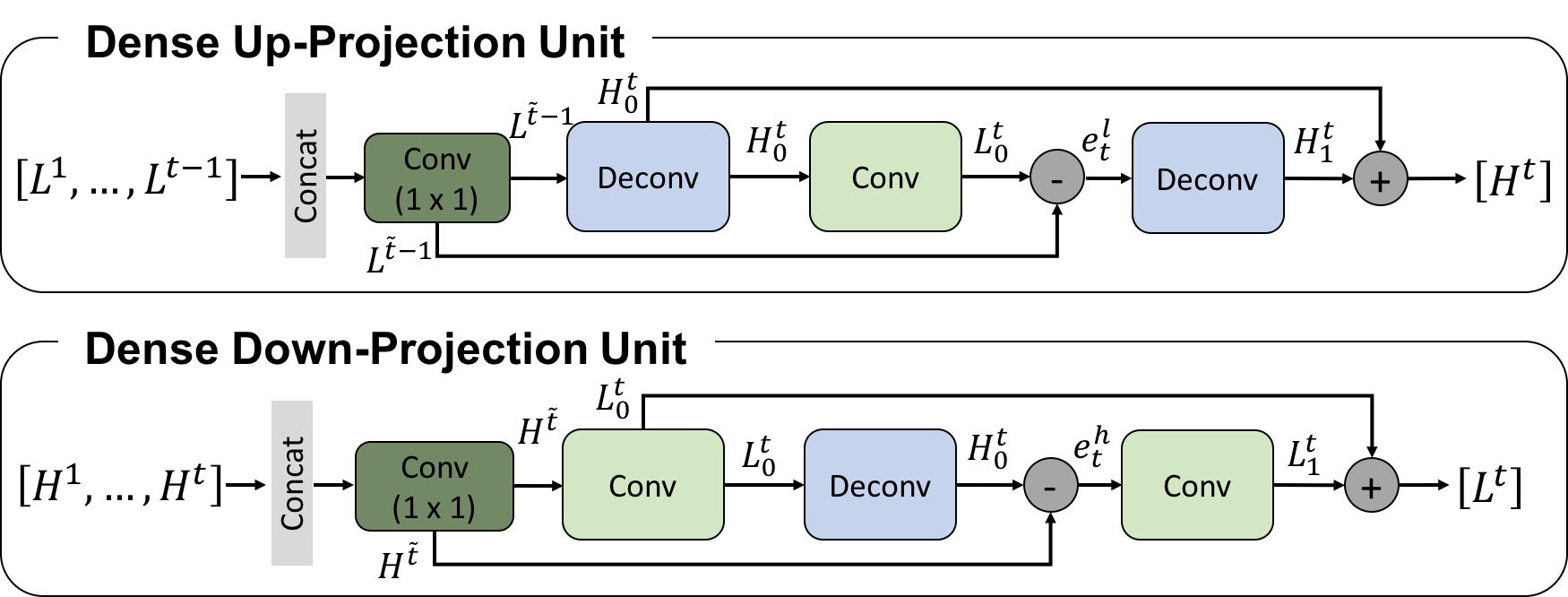}\vspace{-1em}
\caption{Proposed up- and down-projection unit in the Dense DBPN. The feature maps of all preceding units (i.e., $[L^{1}, ..., L^{t-1}]$ and $[H^{1}, ..., H^{t}]$ in up- and down-projections units, respectively) are concatenated and used as inputs, and its own feature maps are used as inputs into all subsequent units.}
\label{figure:D_DBPN}
\end{figure}

In Dense DBPN, the input for each unit is the concatenation of the outputs from all previous units. Let the $L^{\tilde{t}}$ and $H^{\tilde{t}}$ be the input for dense up- and down-projection unit, respectively. They are generated using \texttt{conv}$(1,n_R)$ which is used to merge all previous outputs from each unit as shown in~Fig.~\ref{figure:D_DBPN}. This improvement enables us to generate the feature maps effectively, as shown in the experimental results.

\subsection{Recurrent DBPN}
\label{subsec:recurrent_dbpn}
Here, we propose recurrent DBPN which is able to reduce the number of parameters and widen the receptive field without increasing the model capacity.
In SISR, DRCN~\cite{kim2016deeply} proposed recursive layers without introducing new parameters for additional convolutions in the networks. 
Then, DRRN~\cite{Tai-DRRN-2017} improves residual networks by introducing both global and local residual learning using a very deep CNN model (up to 52-layers).
DBPN can also be treated as a recurrent network by sharing the projection units across the stages. We divided recurrent DBPN into two variants as mentioned below.

\begin{figure}[t!]
\centering
\includegraphics[width=9cm]{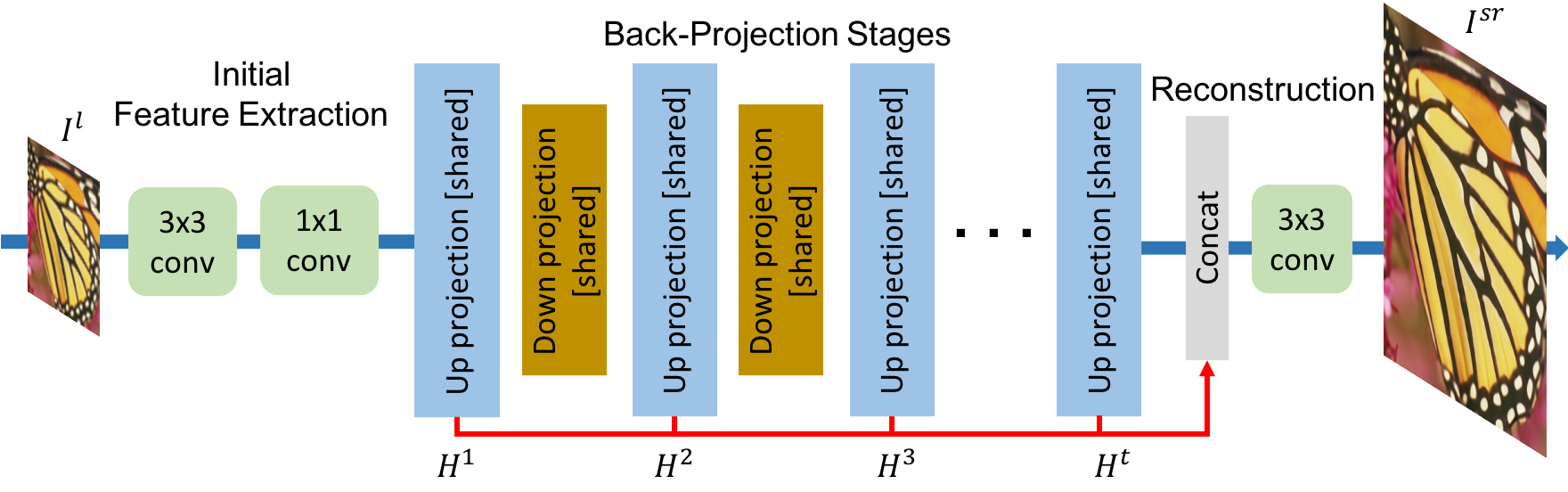}\vspace{-1.5em}
\caption{Recurrent DBPN with shared parameter (DBPN-R).}
\label{figure:dbpnrec}
\end{figure}

\noindent(a) \textbf{Parameter sharing on projection unit (DBPN-R)}. 
This variant uses only one up-projection unit and one down-projection unit which is shared across all stages without dense connection as shown in Fig.~\ref{figure:dbpnrec}. 

\begin{figure}[t!]
\centering
\includegraphics[width=9cm]{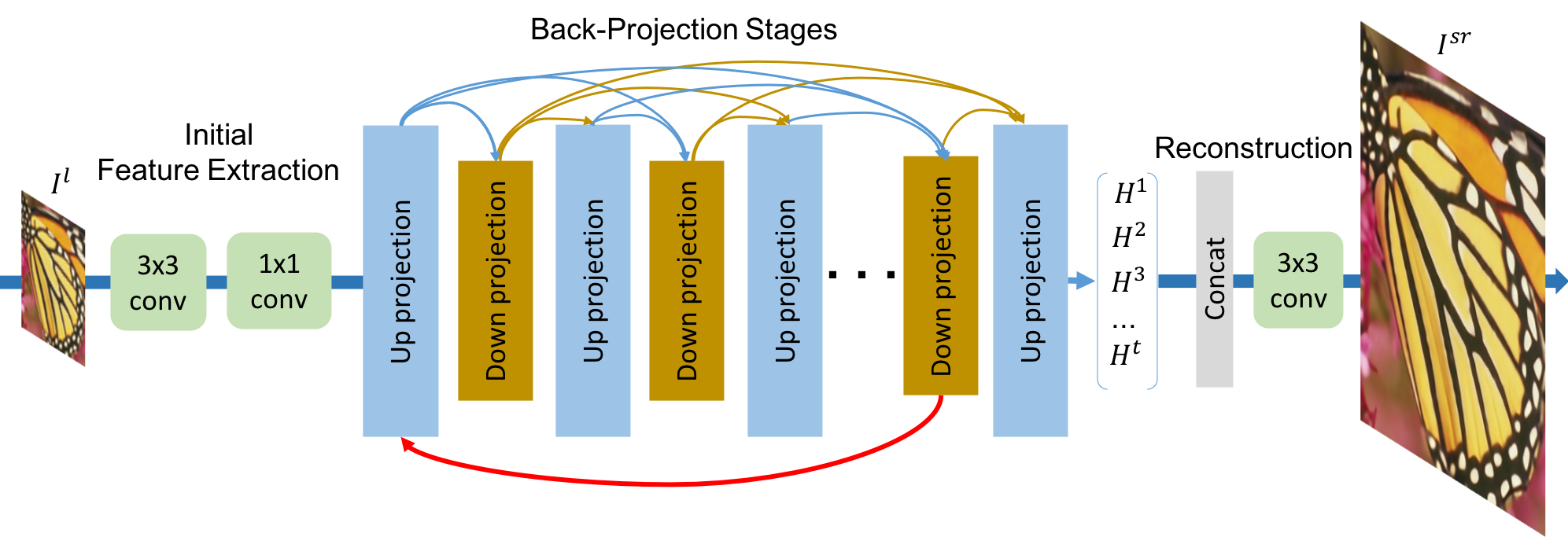}\vspace{-1.5em}
\caption{Recurrent DBPN with transition layer (DBPN-MR).}
\label{figure:dbpniter}
\end{figure}

\noindent(b) \textbf{Transition layer on projection step (DBPN-MR)}. 
This variant uses multiple up- and down-projection units as shown in Fig.~\ref{figure:dbpniter}. 
The last up- and down-projection units were used as transition layer. 
Instead of taking the output from each up-projection unit, DBPN-MR takes the HR features only from the last up-projection unit, then concatenates the HR features from each iteration. 
Here, the output from the last down-projection unit is the input for the first up-projection layer on the next iteration.
Then, the last up-projection unit will receive the output of all previous down-projection units in the corresponding iteration.

\subsection{Residual DBPN}
\label{subsec:residual_dbpn}
Residual learning helps the network to converge faster and make the network have an easier job to produce only the difference between HR and interpolated LR image.
Initially, residual learning has been applied in SR by VDSR~\cite{Kim_2016_VDSR}.
Residual DBPN takes LR image as an input to reduce the computational time.
First, LR image is interpolated using Bicubic interpolation; then, at the last stage, the interpolated image is added to the reconstructed image to produce final SR image.


\section{Experimental Results}
\label{sec:experiment}
\subsection{Implementation and training details}
In the proposed networks, the filter size in the projection unit is
various with respect to the scaling factor. For $2\times$,
we use $6 \times 6$ kernel with stride = 2 and pad by 2 pixels. 
Then, $4\times$ use $8 \times 8$ kernel with stride = 4 and pad by 2 pixels. 
Finally, the $8\times$ use $12\times 12$ kernel with stride = 8 and pad by 2.\footnote{We found these settings to work well based on general intuition and preliminary experiments.}

We initialize the weights based on~\cite{he2015delving}. Here, standard deviation (std) is computed by $(\sqrt{2/n_l})$ where $n_l=f^{2}_{t}n_{t}$, $f_{t}$ is the filter size, and $n_t$ is the number of filters. For example, with $f_{t}=3$ and $n_t=8$, the std is $0.111$. All convolutional and deconvolutional layers are followed by parametric rectified linear units (PReLUs), except the final reconstruction layer.

We trained all networks using images from DIV2K \cite{Agustsson_2017_CVPR_Workshops} with augmentation (scaling, rotating, flipping, and random cropping). To produce LR images, we downscale the HR images on particular scaling factors using Bicubic. We use batch size of 16 with size $40 \times 40$ for LR image, while HR image size corresponds to the scaling factors. The learning rate is initialized to $1e-4$ for all layers and decrease by a factor of 10 for every $5\times 10^5$ iterations for total $10^6$ iterations. We used Adam with momentum to $0.9$ and trained with L1 Loss. All experiments were conducted using PyTorch 0.3.1 and Python 3.5 on NVIDIA TITAN X GPUs. The code is available in the internet.\footnote{The implementation is available \href{https://www.toyota-ti.ac.jp/Lab/Denshi/iim/members/muhammad.haris/projects/DBPN.html}{here}.}

\begin{table*}[t!]
\scriptsize
\caption{Model architecture of DBPN. "Feat0" and "Feat1" refer to first and second convolutional layer in the initial feature extraction stages. Note: ($f,n,st,pd$) where $f$ is filter size, $n$ is number of filters, $st$ is striding, and $pd$ is padding}
\centering
\label{tab:net_arc}
\begin{tabular}{*1l|*1c|*1c*1c*1c*1c*1c*1c*1c}
\hline
& \textbf{Scale} & \textbf{DBPN-SS}&\textbf{DBPN-S} & \textbf{DBPN-M}& \textbf{DBPN-L}& \textbf{D-DBPN-L}& \textbf{D-DBPN} & \textbf{DBPN}   \\[4pt]
\hline
Input/Output&&Luminance&Luminance&Luminance&Luminance&Luminance&RGB&RGB\\[4pt]
\hline
Feat0&&(3,64,1,1)&(3,128,1,1)&(3,128,1,1)&(3,128,1,1)&(3,128,1,1)&(3,256,1,1)&(3,256,1,1)\\[4pt]
\hline
Feat1&&(1,18,1,0)&(1,32,1,0)&(1,32,1,0)&(1,32,1,0)&(1,32,1,0)&(1,64,1,0)&(1,64,1,0)\\[4pt]
\hline
Reconstruction&&(1,1,1,0)&(1,1,1,0)&(1,1,1,0)&(1,1,1,0)&(1,1,1,0)&(3,3,1,1)&(3,3,1,1)\\[4pt]
\hline
&$2\times$&(6,18,2,2)&(6,32,2,2)&(6,32,2,2)&(6,32,2,2)&(6,32,2,2)&(6,64,2,2)&(6,64,2,2)\\[4pt]
\cline{2-9}
BP stages&$4\times$&(8,18,4,2)&(8,32,4,2)&(8,32,4,2)&(8,32,4,2)&(8,32,4,2)&(8,64,4,2)&(8,64,4,2)\\[4pt]
\cline{2-9}
&$8\times$&(12,18,8,2)&(12,32,8,2)&(12,32,8,2)&(12,32,8,2)&(12,32,8,2)&(12,64,8,2)&(12,64,8,2)\\[4pt]
\hline
&$2\times$&106&337&779&1221&1230&5819&8811\\[4pt]
\cline{2-9}
Parameters ($k$) &$4\times$&188&595&1381&2168&2176&10426&15348\\[4pt]
\cline{2-9}
&$8\times$&421&1332&3101&4871&4879&23205&34026\\[4pt]
\hline
Depth&&12&12&24&36&40&52&76\\[4pt]
\hline
No. of stage ($T$)&&2&2&4&6&6&7&10\\[4pt]
\hline
Dense connection&&No&No&No&No&Yes&Yes&Yes\\[4pt]
\hline
\end{tabular}
\end{table*}

\subsection{Model analysis}
There are six types of DBPN used for model analysis: DBPN-SS, DBPN-S, DBPN-M, DBPN-L, D-DBPN-L, D-DBPN, and DBPN.
The detailed architectures of those networks are shown in Table~\ref{tab:net_arc}.
Other methods, VDSR~\cite{Kim_2016_VDSR}, DRCN~\cite{kim2016deeply}, DRRN~\cite{Tai-DRRN-2017}, LapSRN~\cite{LapSRN}, was chosen due to the same nature in number of parameter.

\label{subsec:modelanalysis}
\begin{figure}[t!]
\centering
\includegraphics[width=8.5cm]{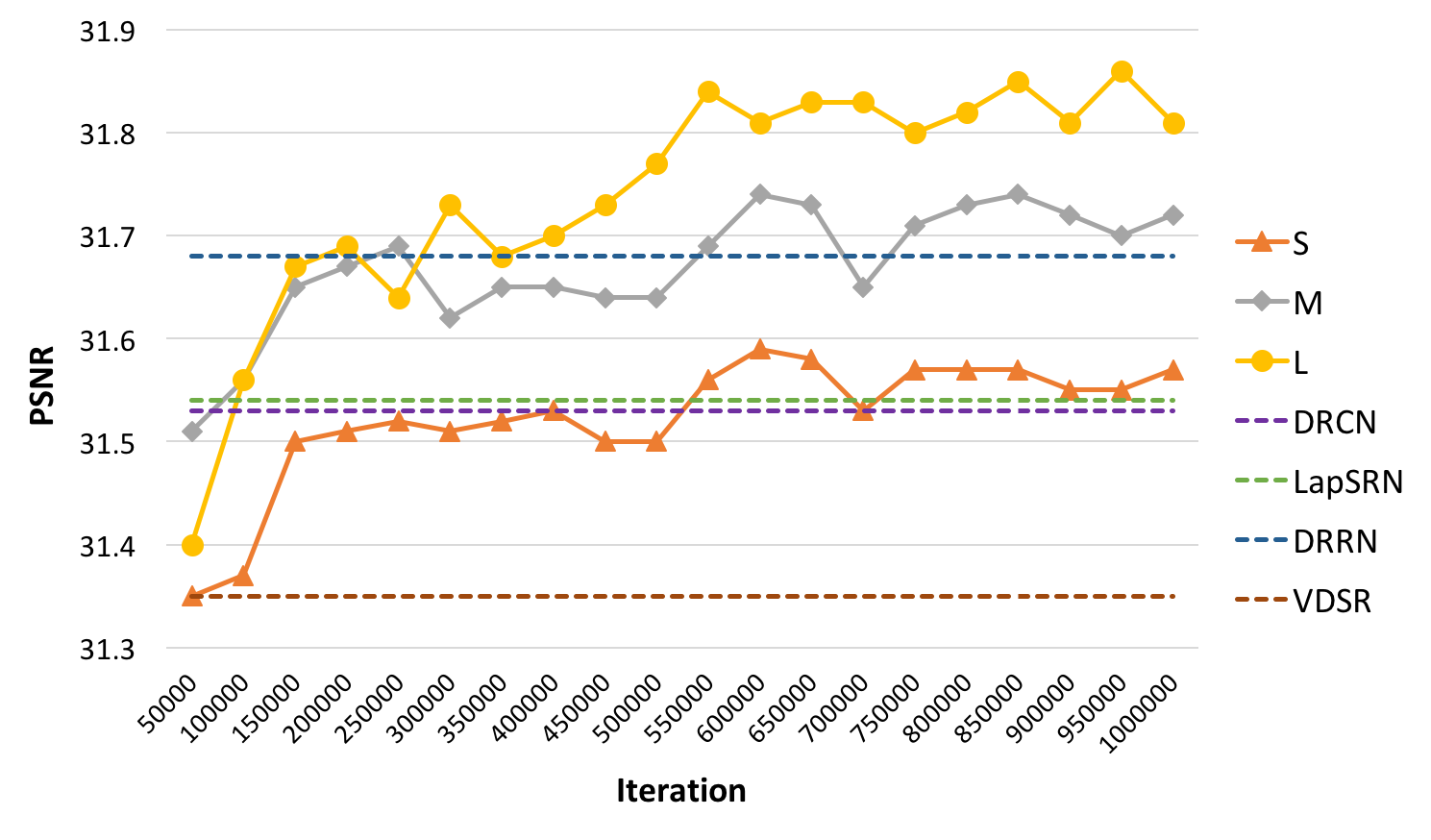}\vspace{-1.5em}
\caption{The depth analysis of DBPNs compare to other networks (VDSR~\cite{Kim_2016_VDSR}, DRCN~\cite{kim2016deeply}, DRRN~\cite{Tai-DRRN-2017}, LapSRN~\cite{LapSRN}) on Set5 dataset for 4$\times$ enlargement.}
\label{figure:modular_comparison_4x}
\end{figure} 

\noindent \textbf{Depth analysis}. To demonstrate the capability of our projection unit, we construct multiple networks: DBPN-S ($T=2$), DBPN-M ($T=4$), and DBPN-L ($T=6$). In the feature extraction, we use $n_{0} = 128$ and $n_{R} = 32$. Then, we use $\texttt{conv}(1,1)$ for the reconstruction. The input and output image are luminance only. 

The results on $4\times$ enlargement are shown in~Fig.~\ref{figure:modular_comparison_4x}. 
From the first $50k$ iteration, our variants are outperformed VDSR.
Finally, starting from our shallow network, DBPN-S gives the higher PSNR than VDSR, DRCN, and LapSRN. 
DBPN-S uses only 12 convolutional layers with smaller number of filters than VDSR, DRCN, and LapSRN. At the best performance, DBPN-S can achieve $31.59$ dB which better $0.24$ dB, $0.06$ dB, $0.05$ dB than VDSR, DRCN, and LapSRN, respectively. DBPN-M shows performance improvement which better than all four existing methods (VDSR, DRCN, LapSRN, and DRRN). At the best performance, DBPN-M can achieve $31.74$ dB which better $0.39$ dB, $0.21$ dB, $0.20$ dB, $0.06$ dB than VDSR, DRCN, LapSRN, and DRRN respectively. In total, DBPN-M uses 24 convolutional layers which has the same depth as LapSRN. Compare to DRRN (up to 52 convolutional layers), DBPN-M undeniable shows the effectiveness of our projection unit. Finally, DBPN-L outperforms all methods with $31.86$ dB which better $0.51$ dB, $0.33$ dB, $0.32$ dB, $0.18$ dB than VDSR, DRCN, LapSRN, and DRRN, respectively.

\begin{figure}[t!]
\centering
\includegraphics[width=8.5cm]{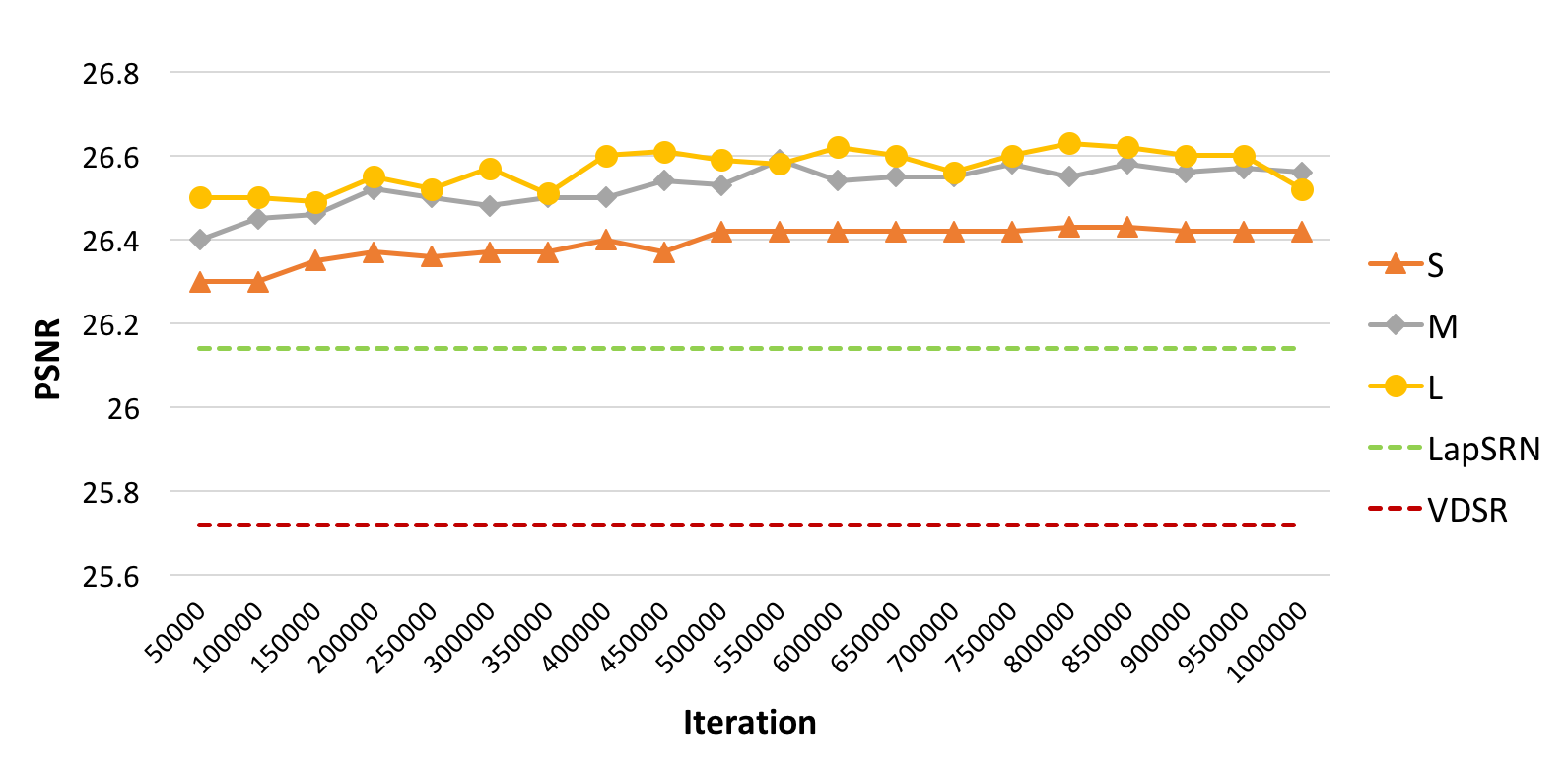}\vspace{-1.5em}
\caption{The depth analysis of DBPN on Set5 dataset for 8$\times$ enlargement. DBPN-S ($T=2$), DBPN-M ($T=4$), and DBPN-L ($T=6$)}
\label{figure:modular_comparison_8x}
\end{figure}

The results of $8\times$ enlargement are shown in~Fig.~\ref{figure:modular_comparison_8x}. 
Our networks outperform the existing networks for $8\times$ enlargement, from the first $50k$ iteration, which clearly show the effectiveness of our proposed networks on large scaling factors. 
However, we found that there is no significant performance gain from each proposed network especially for DBPN-L and DBPN-M networks where the difference only $0.04$ dB. 

\begin{figure}[t!]
\centering
\includegraphics[width=8.5cm]{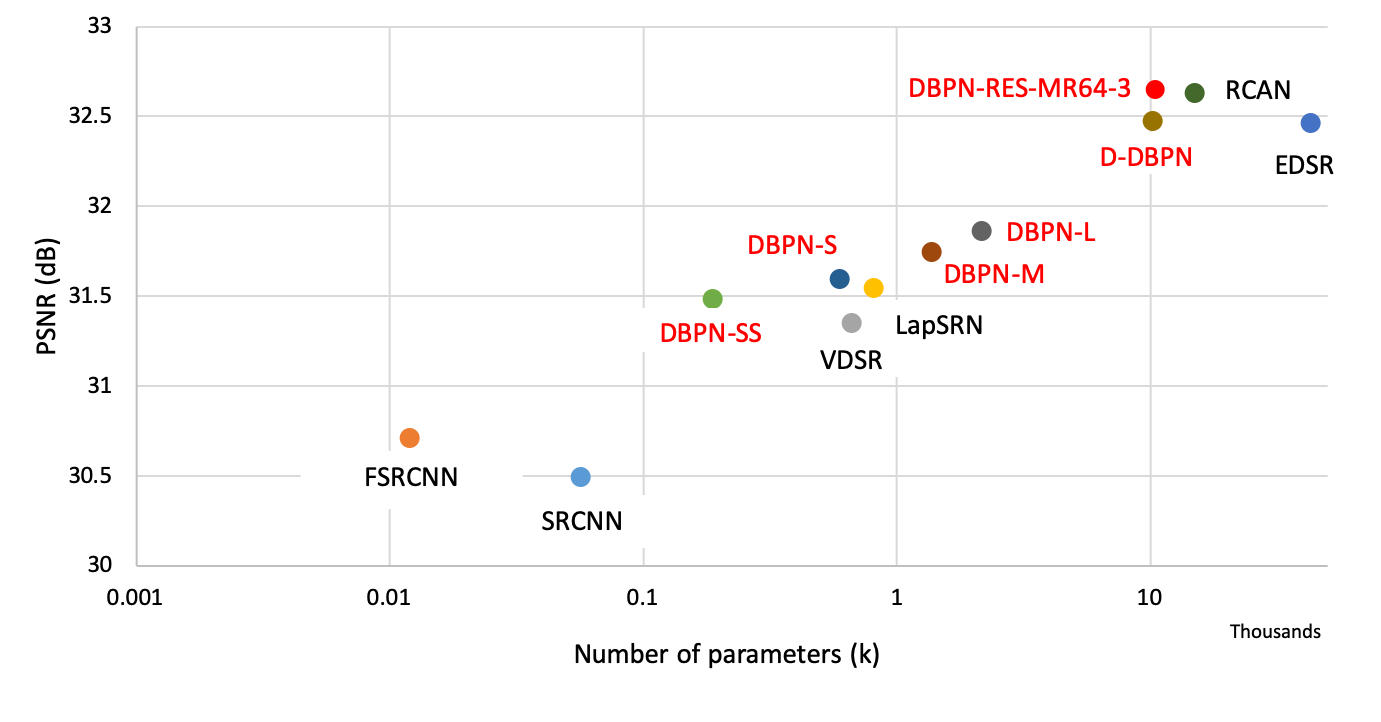}\vspace{-1.5em}
\caption{Performance vs number of parameters for $4\times$ enlargement using Set5. The horizontal axis is log-scale.}
\label{figure:psnr_vs_param_4x}

\includegraphics[width=8.5cm]{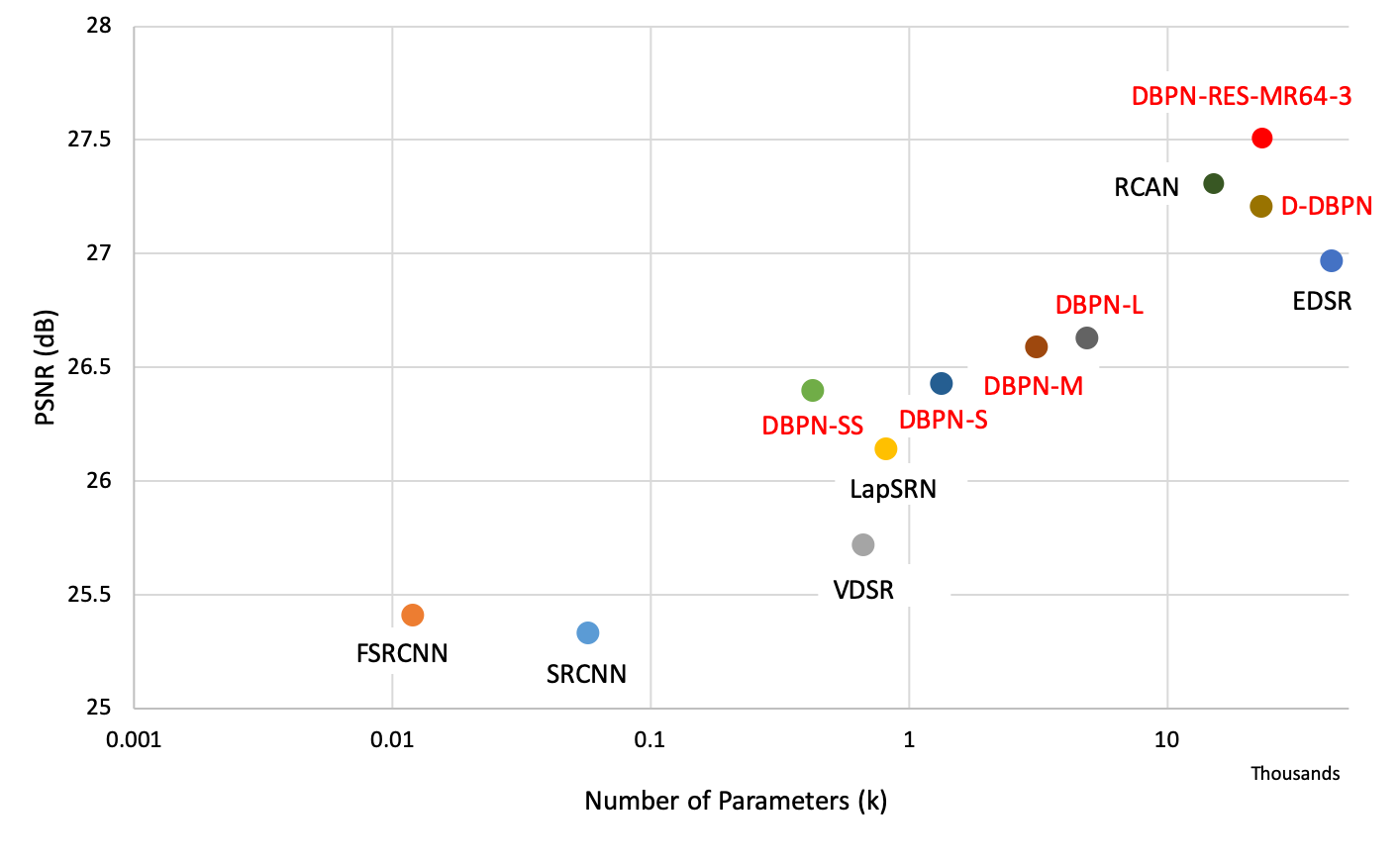}\vspace{-1.5em}
\caption{Performance vs number of parameters for $8\times$ enlargement using Set5. The horizontal axis is log-scale.}
\label{figure:psnr_vs_param_8x}
\end{figure}

\noindent \textbf{Number of parameters}. 
We show the tradeoff between performance
and number of network parameters from our networks and existing deep
network SR in~Fig.~\ref{figure:psnr_vs_param_4x}~and~\ref{figure:psnr_vs_param_8x}. 

For the sake of low computation for real-time processing, we construct DBPN-SS which is the lighter version of DBPN-S, $(T=2)$. We use $n_{0} = 64$ and $n_{R} = 18$. However, the results outperform SRCNN, FSRCNN, and VDSR on both $4\times$ and $8\times$ enlargement. Moreover, DBPN-SS performs better than VDSR with $72\%$ and $37\%$ fewer parameters on $4\times$ and $8\times$ enlargement, respectively. 

DBPN-S has about $27\%$ fewer parameters and higher PSNR than LapSRN on $4\times$ enlargement. 
Finally, D-DBPN has about $76\%$ fewer parameters, and
approximately the same PSNR, compared to EDSR on $4\times$ enlargement. On the $8\times$ enlargement, D-DBPN has about $47\%$ fewer parameters with better PSNR compare to EDSR. This evidence show that our networks has the best trade-off between performance and number of parameter.

%

\begin{figure}[t!]
\centering
\includegraphics[width=8.5cm]{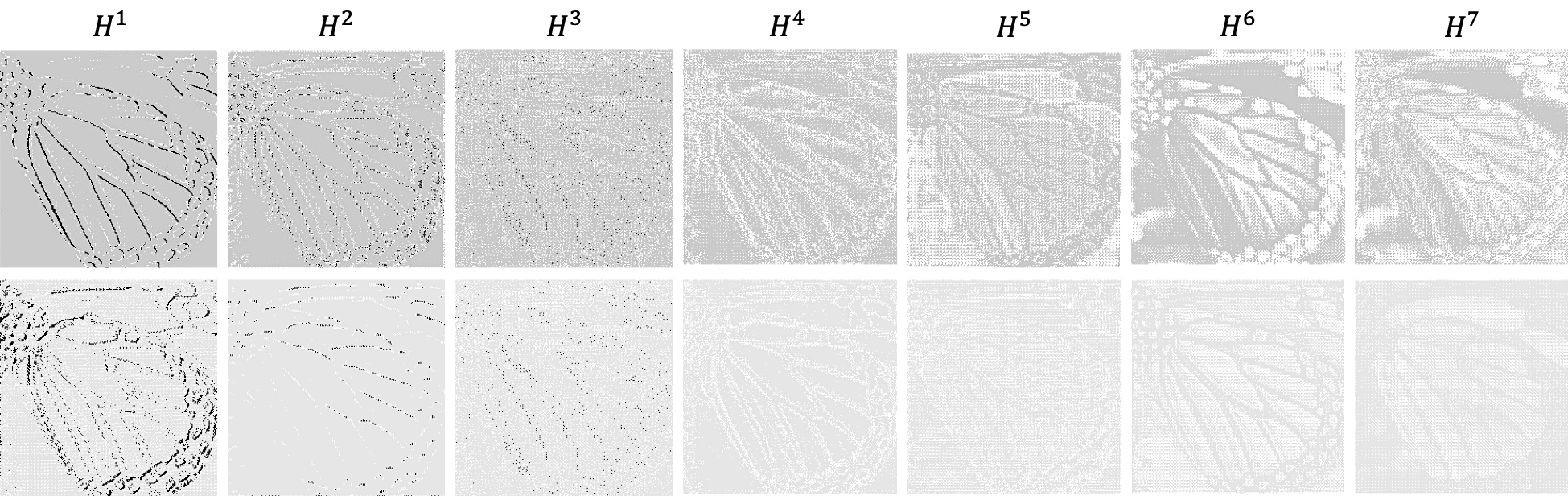}
\caption{Sample of feature maps from up-projection units in D-DBPN where $t=7$. Each feature has been enhanced using the same grayscale colormap. Zoom in for better visibility.}
\label{figure:result_up_projection}
\end{figure}

\noindent \textbf{Deep concatenation}. Each projection unit is used to
distribute the reconstruction step by constructing features
which represent different details of the HR components. Deep concatenation is also well-related with the number of $T$ (back-projection stage), 
which shows more detailed features generated from the projection units will also increase the quality of the results. In~Fig.~\ref{figure:result_up_projection}, it is shown that each stage successfully generates diverse features to reconstruct SR image.

\noindent \textbf{Error Feedback}.
As stated before, error feedback (EF) is used to guide the reconstruction in the early layer. Here, we analyze how error feedback can help for better reconstruction. 
We conduct experiments to see the effectiveness of error feedback procedure.
On the scenario without EF, we replace up- and down-projection unit with single up- (deconvolution) and down-sampling (convolution) layer.

\begin{figure}[t!]
\centering
\includegraphics[width=8.5cm]{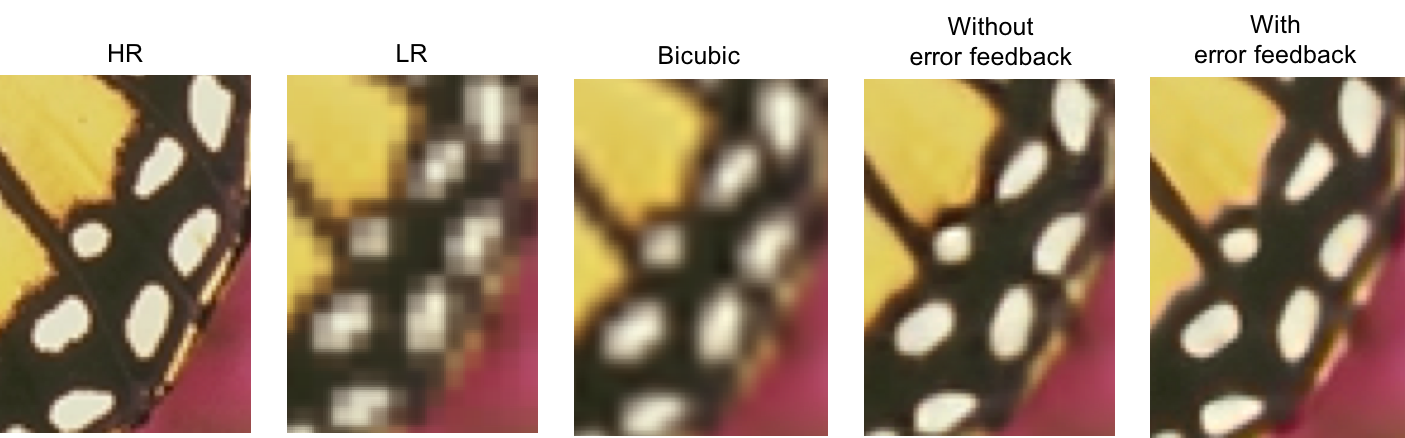}
\caption{Qualitative comparisons of DBPN-S with EF and without EF on $4\times$ enlargement.}
\label{figure:error_feedback}
\end{figure}

We show PSNR of DBPN-S with EF and without EF in~Table~\ref{tab:error_feedback}. The result with EF has 0.53 dB and 0.26 dB better than without EF on Set5 and Set14, respectively. In~Fig.~\ref{figure:error_feedback}, we visually show how error feedback can construct better and sharper HR image especially in the white stripe pattern of the wing.

Moreover, the performance of DBPN-S without EF is interestingly 0.57 dB and 0.35 dB better than previous approaches such as SRCNN~\cite{dong2016image} and FSRCNN~\cite{dong2016accelerating}, respectively, on Set5. The results show the effectiveness of iterative up- and downsampling layers to demonstrate the LR-to-HR mutual dependency.

\begin{table}[t!]
\caption{Analysis of EF using DBPN-S on $4\times$ enlargement. {\color{red}Red} indicates the best performance.}
\centering
\label{tab:error_feedback}
\begin{tabular}{*1l|*1c|*1c}
\hline
& \textbf{Set5} & \textbf{Set14} \\
\hline
SRCNN~\cite{dong2016image}&30.49&27.61\\[4pt]
\hline
FSRCNN~\cite{dong2016accelerating}&30.71&27.70\\[4pt]
\hline
Without EF&31.06&27.95\\[4pt]
\hline
With EF&{\color{red}31.59}&{\color{red}28.21}\\[4pt]
\hline
\end{tabular}
\end{table}

We further analyze the effectiveness of EF comparing the same model size as shown in Table~\ref{tab:bp_analysis} using D-DBPN model. 
The better setting is to remove the subtraction (-) and addition (+) operations in the up-/down-projection unit. The results demonstrate the effectiveness of our EF module.

\begin{table}[t!]
\caption{Analysis of EF module on same model size (D-DBPN) on $4\times$ enlargement. {\color{red}Red} indicates the best performance.}
\centering
\label{tab:bp_analysis}
\begin{tabular}{*1l|*1c|*1c}
\hline
& \textbf{D-DBPN-w/ EF} & \textbf{D-DBPN-w/o EF} \\
\hline
Set5&{\color{red}32.40/0.897}&32.17/0.894\\[4pt]
\hline
Set14&{\color{red}28.75/0.785}&28.56/0/782\\[4pt]
\hline
BSDS100&{\color{red}27.67/0.738}&27.56/0.736\\[4pt]
\hline
Urban100&{\color{red}26.38/0.793}&26.09/0.786\\[4pt]
\hline
Manga109&{\color{red}30.89/0.913}&30.43/0.908\\[4pt]
\hline
\end{tabular}
\end{table}

\noindent \textbf{Filter Size}
We analyze the size of filters which is used in the back-projection stage on D-DBPN model. As stated before, the choice of filter size in the back-projection stage is based on the preliminary results. For the 4$\times$ enlargement, we show that filter 8$\times$8 is 0.08 dB and 0.09 dB better than filter 6$\times$6 and 10$\times$10, respectively, as shown in Table \ref{tab:filter}.
\begin{table}[t!]
\caption{Analysis of filter size in the back-projection stages on $4\times$ enlargement from D-DBPN. {\color{red}Red} indicates the best performance.}
\centering
\label{tab:filter}
\begin{tabular}{*1c|*1c|*1c|*1c|*1c}
\hline
Filter size&Striding&Padding& \textbf{Set5} & \textbf{Set14} \\
\hline
6&4&1&32.39&28.78\\[4pt]
\hline
8&4&2&{\color{red}32.47}&{\color{red}28.82}\\[4pt]
\hline
10&4&3&32.38&28.79\\[4pt]
\hline
\end{tabular}
\end{table}

\noindent \textbf{Luminance vs RGB}
In D-DBPN, we change the input/output from luminance to RGB color channels. 
There is no significant improvement in the quality of the result as shown in~Table~\ref{tab:color}. 
However, for running time efficiency, constructing all channels simultaneously is faster than a separated process.
\begin{table}[t!]
\caption{Analysis of input/output color channel using DBPN-L. {\color{red}Red} indicates the best performance.}
\centering
\label{tab:color}
\begin{tabular}{*1l|*1c|*1c}
\hline
& \textbf{Set5} & \textbf{Set14} \\
\hline
RGB&{\color{red}31.88}&{\color{red}28.47}\\[4pt]
\hline
Luminance&31.86&{\color{red}28.47}\\[4pt]
\hline
\end{tabular}
\end{table}


\subsection{Comparison of each DBPN variant}
\noindent \textbf{Dense connection}. We implement D-DBPN-L which is a dense connection of the $L$ network to show how dense connection can improve the network's performance in all cases as shown in~Table~\ref{tab:dense}. On $4\times$ enlargement, the dense network, D-DBPN-L, gains $0.13$ dB and $0.05$ dB higher than DBPN-L on the Set5 and Set14, respectively. On $8\times$, the gaps are even larger. The D-DBPN-L has $0.23$ dB and $0.19$ dB higher that DBPN-L on the Set5 and Set14, respectively.
\begin{table}[t!]
\small
\caption{Comparison of the DBPN-L and D-DBPN-L on 4$\times$ and 8$\times$ enlargement. {\color{red}Red} indicates the best performance.}
\centering
\label{tab:dense}
\begin{tabular}{*1l*1c|*2c*2c}
\hline
 & &\multicolumn{2}{c}{Set5} & \multicolumn{2}{c}{Set14} \\         
Algorithm & Scale & PSNR&SSIM & PSNR&SSIM   \\
\hline
DBPN-L&4		&$31.86$&$0.891$&$28.47$&$0.777$\\
D-DBPN-L&4		&{\color{red}$31.99$}&{\color{red}$0.893$}&{\color{red}$28.52$}&{\color{red}$0.778$}\\
\hline
DBPN-L&8		&$26.63$&$0.761$&$24.73$&$0.631$\\
D-DBPN-L&8		&{\color{red}$26.86$}&{\color{red}$0.773$}&{\color{red}$24.92$}&{\color{red}$0.638$}\\
\hline
\end{tabular}
\end{table}

\noindent \textbf{Comparison across the variants}. We compare six DBPN variants: DBPN-R64-10, DBPN-R128-5, DBPN-MR64-3, DBPN-RES-MR64-3, DBPN-RES, and DBPN. 
First, DBPN, which was the winner of NTIRE2018~\cite{timofte2018ntire} and PIRM2018~\cite{pirm2018}, uses $n_{0} = 256$, $n_{R} = 64$, and $t=10$ for the back-projection stages, and dense connection between projection units. In the reconstruction, we use $\texttt{conv}(3,3)$. 
DBPN-R64-10 uses $n_R=64$ with 10 iterations to produce 640 HR features as input of reconstruction layer.
DBPN-R128-5, uses $n_R=128$ with 5 iterations, produces 640 HR features.
DBPN-MR64-3 has the same architecture with D-DBPN but the projection units are treated as recurrent network.
DBPN-RES-MR64-3 is DBPN-MR64-3 with residual learning. 
Last, DBPN-RES is DBPN with residual learning. 
All variants are trained with the same training setup.

The results are shown in Table~\ref{tab:psnr_variant}. It shows that all variants successfully have better performance than D-DBPN~\cite{haris2018deep}.
DBPN-R64-10 has the least parameter compare to other variants, which is suitable for mobile/real-time application. It can reduce $10\times$ number of parameter compare to DBPN and maintain to get good performance.
We can see that increasing $n_R$ can improve the performance of DBPN-R which is shown by DBPN-R128-5 compare to DBPN-R64-10.
However, better results is obtained by DBPN-MR64-3, especially on Urban100 and Manga109 test set compare to other variants.
It is also proven that residual learning can slightly improve the performance of DBPN.
Therefore, it is natural that we performed the combination of multiple stages recurrent and residual learning called DBPN-RES-MR64-3 which performs the best results and has lower parameter than DBPN.

\begin{table*}[t!]
\scriptsize
\caption{Quantitative evaluation of DBPN's variants on 4$\times$. {\color{red}Red} indicates the best performance. }
\centering
\label{tab:psnr_variant}
\begin{tabular}{*1l*1r*2c*2c*2c*2c*2c}
\hline
 & &\multicolumn{2}{c}{Set5} & \multicolumn{2}{c}{Set14}& \multicolumn{2}{c}{BSDS100}& \multicolumn{2}{c}{Urban100}&\multicolumn{2}{c}{Manga109} \\         
Method & \# Parameters ($k$) &  PSNR&SSIM & PSNR&SSIM & PSNR&SSIM & PSNR&SSIM & PSNR&SSIM\\
\hline
D-DBPN~\cite{haris2018deep} &10426					&{$32.40$}&{$0.897$}&{$28.75$}&{$0.785$}&{$27.67$}&{$0.738$}&{$26.38$}&{$0.793$}&{$30.89$}&{$0.913$}\\
DBPN &15348					&$32.55$&$0.898$&$28.91$&$0.789$&$27.77$&$0.742$&$26.82$&$0.807$&$31.46$&$0.918$\\
DBPN-R64-10&{\color{red}1614}				&$32.38$&$0.896$&$28.83$&$0.787$&$27.73$&$0.740$&$26.51$&$0.798$&$31.12$&$0.915$\\
DBPN-R128-5&6349				&$32.41$&$0.897$&$28.83$&$0.787$&$27.72$&$0.740$&$26.58$&$0.799$&$31.15$&$0.915$\\
DBPN-MR64-3&10419				&$32.57$&$0.898$&$28.92$&$0.790$&$27.79$&$0.743$&$26.92$&$0.810$&$31.51$&$0.919$\\
DBPN-RES&15348				&$32.54$&$0.897$&$28.92$&$0.789$&$27.79$&$0.742$&$26.89$&$0.808$&$31.49$&$0.918$\\
DBPN-RES-MR64-3&10419				&{\color{red}$32.65$}&{\color{red}$0.899$}&{\color{red}$29.03$}&{\color{red}$0.791$}&{\color{red}$27.82$}&{\color{red}$0.744$}&{\color{red}$27.08$}&{\color{red}$0.814$}&{\color{red}$31.74$}&{\color{red}$0.921$}\\
\hline
\end{tabular}
\end{table*}

\subsection{Comparison with the-state-of-the-arts on SR}
\label{subsec:sota}
\begin{figure*}[t!]
\centering
\includegraphics[width=15cm]{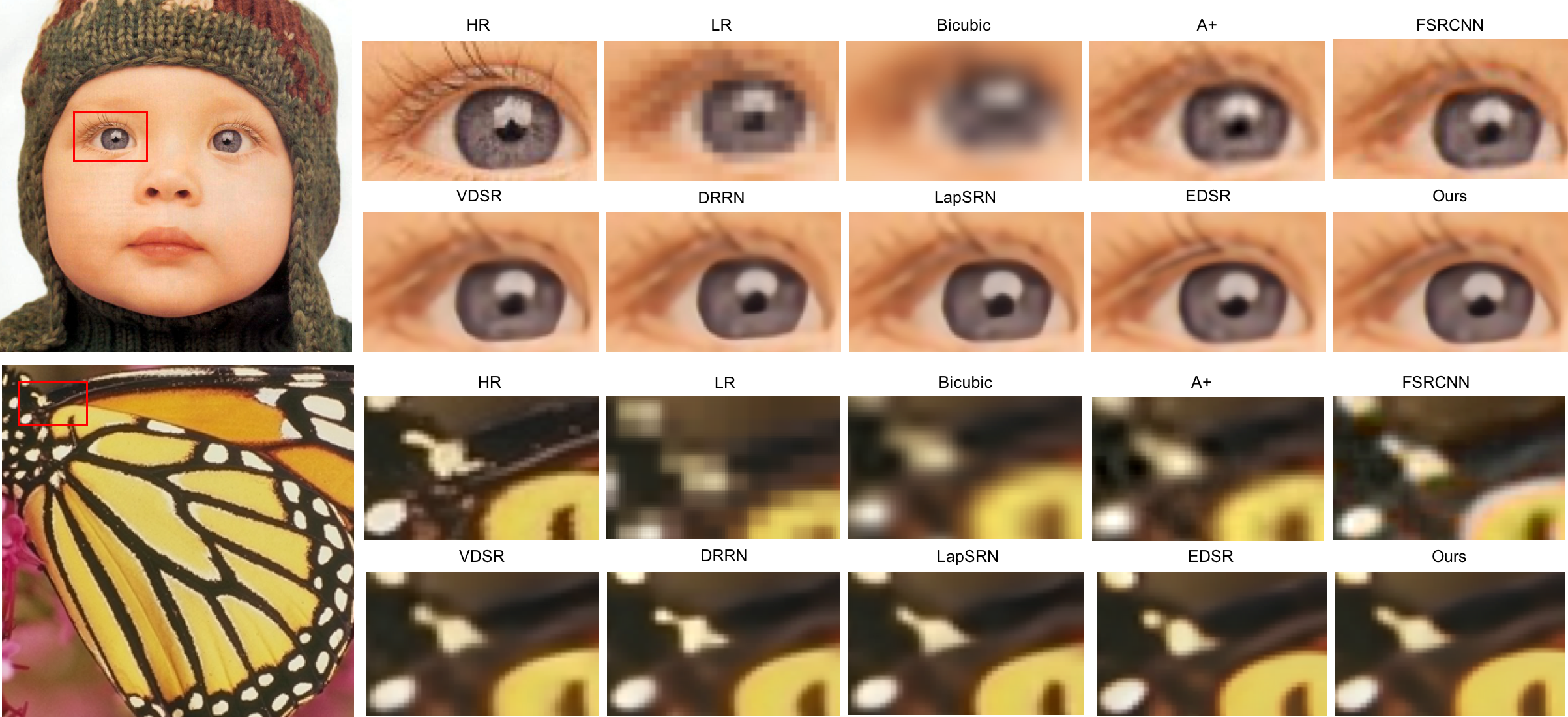}
\caption{Qualitative comparison of our models with other works on $4\times$ super-resolution.}
\label{figure:4x_result}
\end{figure*}

To confirm the ability of the proposed network, we performed several experiments and analysis. We compare our network with 14 state-of-the-art SR algorithms: A+~\cite{timofte2014a+}, SRCNN~\cite{dong2016image}, FSRCNN~\cite{dong2016accelerating}, VDSR~\cite{Kim_2016_VDSR}, DRCN~\cite{kim2016deeply}, DRRN~\cite{Tai-DRRN-2017}, 
LapSRN~\cite{LapSRN}, MS-LapSRN~\cite{lai2018fast}, MSRN~\cite{li2018multi}, D-DBPN~\cite{haris2018deep}, EDSR~\cite{Lim_2017_CVPR_Workshops}, RDN~\cite{zhang2018residual}, RCAN~\cite{zhang2018image}, and SAN~\cite{dai2019second}. We carry out extensive experiments using 5 datasets: Set5~\cite{bevilacqua2012low}, Set14~\cite{zeyde2012single}, BSDS100~\cite{arbelaez2011contour}, Urban100~\cite{huang2015single} and Manga109~\cite{matsui2016sketch}. Each dataset has different characteristics. Set5, Set14 and BSDS100 consist of natural scenes; Urban100 contains urban scenes with details in different frequency bands; and Manga109 is a dataset of Japanese manga. 

Our final network, DBPN-RES-MR64-3, combines dense connection, recurrent network and residual learning to boost the performance of DBPN. It uses $n_{0} = 256$, $n_{R} = 64$, and $t=7$ with 3 iteration. In the reconstruction, we use $\texttt{conv}(3,3)$. RGB color channels are used for input and output image. It takes around 14 days to train\footnote{It takes around five days to train on PyTorch 1.0 and CUDA10.}.

PSNR and structural similarity (SSIM)~\cite{wang04}
were used to quantitatively evaluate the proposed method. Note that higher PSNR
and SSIM values indicate better quality. As used by existing networks,
all measurements used only the luminance channel (Y). For SR by factor
$s$, we crop $s$ pixels near image boundary before evaluation as in~\cite{Lim_2017_CVPR_Workshops, dong2016accelerating}. Some of the existing networks such as SRCNN, FSRCNN, VDSR, and EDSR did not perform $8\times$ enlargement. To this end, we retrained the existing networks by using author's code with the recommended parameters. 

Figure~\ref{figure:4x_result} shows that EDSR tends to generate stronger edge than the ground truth and lead to misleading information in several cases. The result of EDSR shows the eyelashes were interpreted as a stripe pattern. Our result generates softer patterns which is subjectively closer to the ground truth. On the butterfly image, EDSR separates the white pattern and tends to construct regular pattern such ac circle and stripe, while D-DBPN constructs the same pattern as the ground truth. 

We show the quantitative results in the~Table~\ref{tab:psnr}. Our network outperforms the existing methods by a large margin in all scales except RCAN and SAN on $2\times$. 
For $4\times$, EDSR has $0.26$ dB higher than D-DBPN but outperformed by DBPN-RES-MR64-3 with $0.44$ dB margin on Urban100. Recent state-of-the-art, SAN~\cite{dai2019second} and RCAN~\cite{zhang2018image}, performs better results than our network on $2\times$. 
However, on $4\times$, our network has $0.26$ dB higher than RCAN on Urban100. The biggest gap is shown on Manga109, our network has $0.52$ dB higher than RCAN.

Our network shows its effectiveness on $8\times$ enlargement which outperforms all of the existing methods by a large margin. Interesting results are shown on Manga109 dataset where D-DBPN obtains $25.50$ dB which is $0.61$ dB better than EDSR. 
While on the Urban100 dataset, D-DBPN achieves 23.25 which is only $0.13$ dB better than EDSR. 
Our final network, DBPN-RES-MR64-3, outperforms all previous networks. 
DBPN-RES-MR64-3 is roughly $0.2$ dB better than RCAN~\cite{zhang2018image} across multiple dataset. The biggest gap is on Manga109 where DBPN-RES-MR64-3 is $0.47$ dB better than RCAN~\cite{zhang2018image}.
The overall results show that our networks perform better on fine-structures images especially manga characters, even though we do not use any animation images in the training.

\begin{figure*}[t!]
\centering
\includegraphics[width=14.5cm]{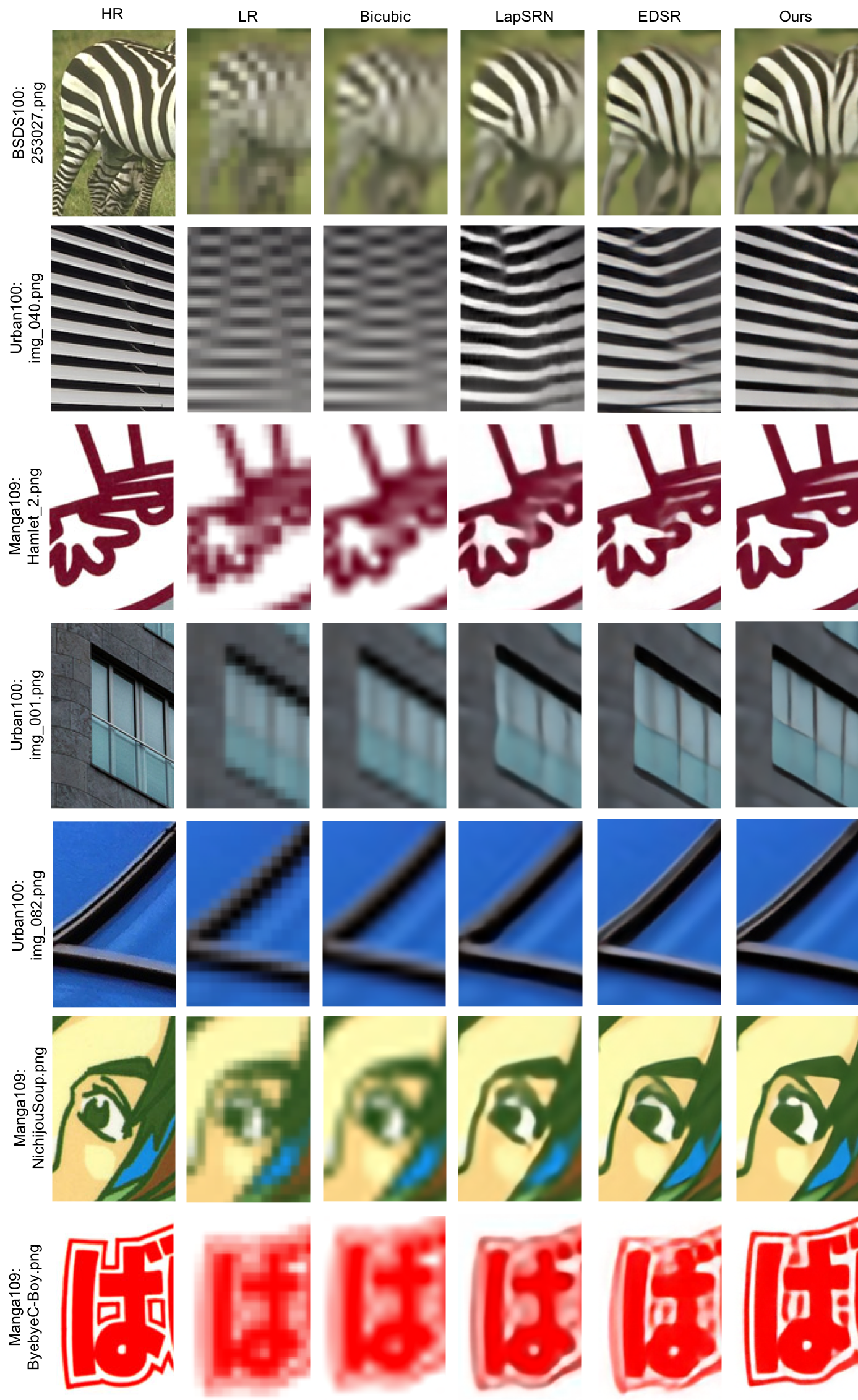}
\caption{Qualitative comparison of our models with other works on $8\times$ super-resolution.}
\label{figure:8x_result}
\end{figure*}

The results of $8\times$ enlargement are visually shown in~Fig.~\ref{figure:8x_result}. Qualitatively, our network is able to preserve the HR components better than other networks. 
For image ``img$\_040$.png'', all of previous methods fail to recover the correct direction of the image textures, while ours produce more faithful results to the ground truth.
For image ``Hamlet$\_2$.png'', other methods suffer from heavy blurring artifacts and fail to recover the details. While, our network successfully recovers the fined detail and produce the closest result to the ground truth.
It shows that our networks can successfully extract not only features but also create contextual information from the LR input to generate HR components in the case of large scaling factors, such as $8\times$ enlargement.


\begin{table*}[t!]
\scriptsize
\caption{Quantitative evaluation of state-of-the-art SR algorithms: average PSNR/SSIM for scale factors 2$\times$, 4$\times$, and 8$\times$. {\color{red}Red} indicates the best and {\color{blue}blue} indicates the second best performance. }
\centering
\label{tab:psnr}
\begin{tabular}{*1l*1c*1c*2c*2c*2c*2c*2c}
\hline
 & & &\multicolumn{2}{c}{Set5} & \multicolumn{2}{c}{Set14}& \multicolumn{2}{c}{BSDS100}& \multicolumn{2}{c}{Urban100}&\multicolumn{2}{c}{Manga109} \\         
Method & Scale &\# Parameters ($M$) & PSNR&SSIM & PSNR&SSIM & PSNR&SSIM & PSNR&SSIM & PSNR&SSIM  \\
\hline
Bicubic&2								&-&$33.65$&$0.930$&$30.34$&$0.870$&$29.56$&$0.844$&$26.88$&$0.841$&$30.84$&$0.935$\\
A+~\cite{timofte2014a+}&2				&-&$36.54$&$0.954$&$32.40$&$0.906$&$31.22$&$0.887$&$29.23$&$0.894$&$35.33$&$0.967$\\
SRCNN~\cite{dong2016image}&2			&0.05M&$36.65$&$0.954$&$32.29$&$0.903$&$31.36$&$0.888$&$29.52$&$0.895$&$35.72$&$0.968$\\
FSRCNN~\cite{dong2016accelerating}&2		&0.01M&$36.99$&$0.955$&$32.73$&$0.909$&$31.51$&$0.891$&$29.87$&$0.901$&$36.62$&$0.971$\\
VDSR~\cite{Kim_2016_VDSR}&2			&0.66M&$37.53$&$0.958$&$32.97$&$0.913$&$31.90$&$0.896$&$30.77$&$0.914$&$37.16$&$0.974$\\
DRCN~\cite{kim2016deeply}&2				&1.77M&$37.63$&$0.959$&$32.98$&$0.913$&$31.85$&$0.894$&$30.76$&$0.913$&$37.57$&$0.973$\\
DRRN~\cite{Tai-DRRN-2017}&2			&0.30M&$37.74$&$0.959$&$33.23$&$0.913$&$32.05$&$0.897$&$31.23$&$0.919$&$37.92$&$0.976$\\
LapSRN~\cite{LapSRN}&2				&0.81M&$37.52$&$0.959$&$33.08$&$0.913$&$31.80$&$0.895$&$30.41$&$0.910$&$37.27 $&$0.974$\\
MS-LapSRN~\cite{lai2018fast}&2			&0.22M&$37.78$&$0.960$&$33.28$&$0.915$&$32.05$&$0.898$&$31.15$&$0.919$&$37.78 $&$0.976$\\
MSRN~\cite{li2018multi}&2				&6.3M&$38.08$&$0.960$&$33.74$&$0.917$&$32.23$&$0.901$&$32.22$&$0.932$&$38.82 $&$0.987$\\
D-DBPN~\cite{haris2018deep} &2			&5.82M&$38.05$&$0.960$&$33.79$&$0.919$&$32.25$&$0.900$&$32.51$&$0.932$&$38.81$&$0.976$\\
RDN~\cite{zhang2018residual}&2			&22.3M&$38.24$&{\color{blue}$0.961$}&$34.01$&{\color{red}$0.921$}&$32.34$&{\color{blue}$0.902$}&$32.89$&$0.935$&$39.18$&{\color{blue}$0.978$}\\
RCAN~\cite{zhang2018image}&2			&16M&{\color{blue}$38.27$}&{\color{blue}$0.961$}&{\color{red}$34.12$}&{\color{red}$0.921$}&{\color{blue}$32.41$}&{\color{red}$0.903$}&{\color{red}$33.34$}&{\color{red}$0.938$}&{\color{red}$39.44$}&{\color{red}$0.979$}\\
SAN~\cite{dai2019second}&2				&15.7M&{\color{red}$38.31$}&{\color{red}$0.962$}&$34.07$&{\color{red}$0.921$}&{\color{red}$32.42$}&{\color{red}$0.903$}&{\color{blue}$33.10$}&{\color{blue}$0.937$}&{\color{blue}$39.32$}&{\color{red}$0.979$}\\
DBPN-RES-MR64-3 &2					&5.82M&$38.08$&$0.960$&{\color{blue}$34.09$}&{\color{red}$0.921$}&$32.31$&$0.901$&$32.92$&$0.935$&$39.28$&$0.977$\\
\hline
Bicubic&4								&-&$28.42$&$0.810$&$26.10$&$0.704$&$25.96$&$0.669$&$23.15$&$0.659$&$24.92$&$0.789$\\
A+~\cite{timofte2014a+}&4				&-&$30.30$&$0.859$&$27.43$&$0.752$&$26.82$&$0.710$&$24.34$&$0.720$&$27.02$&$0.850$\\
SRCNN~\cite{dong2016image}&4			&0.05M&$30.49$&$0.862$&$27.61$&$0.754$&$26.91$&$0.712$&$24.53$&$0.724$&$27.66$&$0.858$\\
FSRCNN~\cite{dong2016accelerating}&4		&0.01M&$30.71$&$0.865$&$27.70$&$0.756$&$26.97$&$0.714$&$24.61$&$0.727$&$27.89$&$0.859$\\
VDSR~\cite{Kim_2016_VDSR}&4			&0.66M&$31.35$&$0.882$&$28.03$&$0.770$&$27.29$&$0.726$&$25.18$&$0.753$&$28.82$&$0.886$\\
DRCN~\cite{kim2016deeply}&4				&1.77M&$31.53$&$0.884$&$28.04$&$0.770$&$27.24$&$0.724$&$25.14$&$0.752$&$28.97$&$0.886$\\
DRRN~\cite{Tai-DRRN-2017}&4			&0.30M&$31.68$&$0.888$&$28.21$&$0.772$&$27.38$&$0.728$&$25.44$&$0.764$&$29.46$&$0.896$\\
LapSRN~\cite{LapSRN}&4				&0.81M&$31.54$&$0.885$&$28.19$&$0.772$&$27.32$&$0.728$&$25.21$&$0.756$&$29.09$&$0.890$\\
MS-LapSRN~\cite{lai2018fast}&4			&0.22M&$31.74$&$0.889$&$28.26$&$0.774$&$27.43$&$0.731$&$25.51$&$0.768$&$29.54$&$0.897$\\
MSRN~\cite{li2018multi}&4				&6.3M&$32.07$&$0.890$&$28.60$&$0.775$&$27.52$&$0.727$&$26.04$&$0.789$&$30.17$&$0.903$\\
D-DBPN~\cite{haris2018deep} &4			&10.2M&$32.40$&$0.897$&$28.75$&$0.785$&$27.67$&$0.738$&$26.38$&$0.793$&$30.89$&$0.913$\\
EDSR~\cite{Lim_2017_CVPR_Workshops}&4	&43.2M&$32.46$&$0.897$&$28.80$&$0.788$&$27.71$&$0.742$&$26.64$&$0.803$&$31.02$&$0.915$\\
RDN~\cite{zhang2018residual}&4			&22.3M&$32.47$&{\color{blue}$0.899$}&$28.81$&$0.787$&$27.72$&$0.742$&$26.61$&$0.803$&$31.00$&$0.915$\\
RCAN~\cite{zhang2018image}&4			&16M&$32.63$&{\color{red}$0.900$}&$28.87$&{\color{blue}$0.789$}&$27.77$&{\color{blue}$0.744$}&{\color{blue}$26.82$}&{\color{blue}$0.809$}&{\color{blue}$31.22$}&{\color{blue}$0.917$}\\
SAN~\cite{dai2019second}&4				&15.7M&{\color{blue}$32.64$}&{\color{red}$0.900$}&{\color{blue}$28.92$}&{\color{blue}$0.789$}&{\color{blue}$27.78$}&$0.743$&$26.79$&$0.807$&$31.18$&{\color{blue}$0.917$}\\
DBPN-RES-MR64-3 &4					&10.2M&{\color{red}$32.65$}&{\color{blue}$0.899$}&{\color{red}$29.03$}&{\color{red}$0.791$}&{\color{red}$27.82$}&{\color{red}$0.744$}&{\color{red}$27.08$}&{\color{red}$0.814$}&{\color{red}$31.74$}&{\color{red}$0.921$}\\
\hline
Bicubic&8								&-&$24.39$&$0.657$&$23.19$&$0.568$&$23.67$&$0.547$&$20.74$&$0.516$&$21.47$&$0.647$\\
A+~\cite{timofte2014a+}&8				&-&$25.52$&$0.692$&$23.98$&$0.597$&$24.20$&$0.568$&$21.37$&$0.545$&$22.39$&$0.680$\\
SRCNN~\cite{dong2016image}&8			&0.05M&$25.33$&$0.689$&$23.85$&$0.593$&$24.13$&$0.565$&$21.29$&$0.543$&$22.37$&$0.682$\\
FSRCNN~\cite{dong2016accelerating}&8		&0.01M&$25.41$&$0.682$&$23.93$&$0.592$&$24.21$&$0.567$&$21.32$&$0.537$&$22.39$&$0.672$\\
VDSR~\cite{Kim_2016_VDSR}&8			&0.66M&$25.72$&$0.711$&$24.21$&$0.609$&$24.37$&$0.576$&$21.54$&$0.560$&$22.83$&$0.707$\\
LapSRN~\cite{LapSRN}&8				&0.81M&$26.14$&$0.738$&$24.44$&$0.623$&$24.54$&$0.586$&$21.81$&$0.582$&$23.39$&$0.735$\\
MS-LapSRN~\cite{lai2018fast}&8			&0.22M&$26.34$&$0.753$&$24.57$&$0.629$&$24.65$&$0.592$&$22.06$&$0.598$&$23.90$&$0.759$\\
MSRN~\cite{li2018multi} &8				&6.3M&$26.59$&$0.725$&$24.88$&$0.5961$&$24.70$&$0.541$&$22.37$&$0.598$&$24.28$&$0.752$\\
D-DBPN~\cite{haris2018deep} &8			&23.1M&$27.25$&$0.785$&$25.14$&$0.649$&$24.91$&$0.602$&$22.72$&$0.630$&$25.14$&$0.798$\\
EDSR~\cite{Lim_2017_CVPR_Workshops}&8	&43.2M&$26.97$&$0.775$&$24.94$&$0.640$&$24.80$&$0.596$&$22.47$&$0.620$&$ 24.58$&$0.778$\\
RCAN~\cite{zhang2018image}&8			&16M&{\color{blue}$27.31$}&{\color{blue}$0.787$}&{\color{blue}$25.23$}&{\color{blue}$0.651$}&{\color{blue}$24.98$}&{\color{blue}$0.606$}&{\color{blue}$23.00$}&{\color{blue}$0.645$}&{\color{blue}$ 25.24$}&{\color{blue}$0.803$}\\
SAN~\cite{dai2019second}&8				&15.7M&$27.22$&$0.783$&$25.14$&$0.648$&$24.88$&$0.601$&$22.70$&$0.631$&$ 24.85$&$0.791$\\
DBPN-RES-MR64-3 &8					&23.1M&{\color{red}$27.51$}&{\color{red}$0.793$}&{\color{red}$25.41$}&{\color{red}$0.657$}&{\color{red}$25.05$}&{\color{red}$0.607$}&{\color{red}$23.20$}&{\color{red}$0.652$}&{\color{red}$25.71$}&{\color{red}$0.813$}\\
\hline
\end{tabular}
\end{table*}

\subsection{Runtime Evaluation}
We present the runtime comparisons between our networks and three existing methods: VDSR~\cite{Kim_2016_VDSR}, DRRN~\cite{Tai-DRRN-2017}, and EDSR~\cite{Lim_2017_CVPR_Workshops}. The comparison must be done in fair settings. The runtime is calculated using python function \texttt{timeit} which encapsulating only \texttt{forward} function. For EDSR, we use original author code based on Torch and use \texttt{timer} function to obtain the runtime.

We evaluate each network using NVIDIA TITAN X GPU (12G Memory). The input image size is 64$\times$ 64, then upscaled into 128$\times$ 128 (2$\times$), 256$\times$ 256 (4$\times$), and 512$\times$ 512 (8$\times$). The results are the average of 10 times trials.

Table~\ref{tab:runtime} shows the runtime comparisons on 2$\times$, 4$\times$, and 8$\times$ enlargement. It shows that DBPN-SS and DBPN-S obtain the best and second best performance on 2$\times$, 4$\times$, and 8$\times$ enlargement. Compare to EDSR, D-DBPN shows its effectiveness by having faster runtime with comparable quality on 2$\times$ and 4$\times$ enlargement. On 8$\times$ enlargement, the gap is bigger. It shows that D-DBPN has better results with lower runtime than EDSR.

Noted that input for VDSR and DRRN is only luminance channel and need preprocessing to create middle-resolution image. So that, the runtime should be added by additional computation of interpolation computation on preprocessing.

\begin{table}[t!]
\caption{Runtime evaluation with input size 64$\times$64. {\color{red}Red} indicates the best and {\color{blue}blue} indicates the second best performance, * indicates the calculation using function timer in Torch, and N.A. indicates that the algorithm runs out of GPU memory.}
\centering
\label{tab:runtime}
\begin{tabular}{*1c|*1c|*1c|*1c}
\hline
& \textbf{$2\times$} & \textbf{$4\times$}& \textbf{$8\times$} \\
& (128$\times$128) & (256$\times$256) & (512$\times$512) \\
\hline
VDSR~\cite{Kim_2016_VDSR}&0.022&0.032&0.068\\[4pt]
\hline
DRRN~\cite{Tai-DRRN-2017}&0.254&0.328&N.A.\\[4pt]
\hline
*EDSR~\cite{Lim_2017_CVPR_Workshops}&0.857&1.245&1.147\\[4pt]
\hline
DBPN-SS&{\color{red}0.012}&{\color{red}0.016}&{\color{red}0.026}\\[4pt]
\hline
DBPN-S&{\color{blue}0.013}&{\color{blue}0.020}&{\color{blue}0.038}\\[4pt]
\hline
DBPN-M&0.023&0.045&0.081\\[4pt]
\hline
DBPN-L&0.035&0.069&0.126\\[4pt]
\hline
D-DBPN&0.153&0.193&0.318\\[4pt]
\hline
DBPN-RES-MR64-3&0.171&0.227&0.339\\[4pt]
\hline
\end{tabular}
\end{table}
\section{Perceptually optimized DBPN}
\label{sec:perceptual}
We also can extend DBPN to produce HR outputs that appear to be better
under human perception. Despite many attempts, it remains unclear how
to accurately model perceptual quality. Instead, we incorporate the
perceptual quality into the generator by using \emph{adversarial
  loss}, as introduced elsewhere~\cite{ledig2016photo, sajjadi2016enhancenet, wang2018recovering}.
In the adversarial settings, there are two building blocks: a
generator ($G$) and a discriminator ($D$). In the context of SR, the
generator $G$ produces HR images (from LR inputs). The discriminator
$D$ works to differentiate between real HR images and generated HR
images (the product of SR network $G$). In our experiments, the
generator is a DBPN network, and the discriminator is a network with
five hidden layers with batch norm, followed by the last, fully connected layer.

\begin{table*}[t]
	\scriptsize
	\caption{\textbf{PIRM2018 Challenge results~\cite{pirm2018}.} The top 9 submissions in each region. For submissions with a marginal PI difference (up to 0.01), the one with the lower RMSE is ranked higher. Submission with marginal differences in both the PI and RMSE are ranked together (marked by $*$).}
	\centering
	\begin{tabu}{@{}llcccllcccllcc@{}}\toprule
		\multicolumn{4}{c}{\textbf{Region 1}} & \phantom{abc}& \multicolumn{4}{c}{\textbf{Region 2}} & \phantom{abc} & \multicolumn{4}{c}{\textbf{Region 3}}\\
		\cmidrule{1	- 4} \cmidrule{6 - 9} \cmidrule{11 - 14} \# & Team & \makecell{PI} & RMSE && \# & Team & \makecell{PI} & RMSE && \# &Team & \makecell{PI} & RMSE\\
		\cmidrule{1	- 4} \cmidrule{6 - 9} \cmidrule{11 - 14} 
		\rowfont{\color{blue}}	$1$  & IPCV \cite{vasu2018analyzing} & 2.709 & 11.48 && $1$ 	& TTI (Ours)			& 2.199	& 12.40	&& $1$	& SuperSR \cite{wang2018esrgan}		& 1.978 & 15.30\\
		\rowfont{\color{blue}}	$2$  & MCML \cite{cheon2018generative} & 2.750	& 11.44	&& $2*$ & IPCV \cite{vasu2018analyzing}		& 2.275 & 12.47	&& $2$  & BOE \cite{Navarrete2018multi}			& 2.019	& 14.24\\
		\rowfont{\color{blue}}	$3*$ & SuperSR \cite{wang2018esrgan} & 2.933	& 11.50	&& $2*$ & MCML \cite{choi2018deep} & 2.279 & 12.41 && $3$  & IPCV \cite{vasu2018analyzing}		& 2.013	& 15.26\\
		\color{blue}{$3*$} & \color{blue}{TTI (Ours)} & \color{blue}{2.938} & \color{blue}{11.46} && $4$ & SuperSR	\cite{wang2018esrgan} & 2.424	& 12.50	&& $4$  & AIM \cite{vu2018perception} & 2.013 & 15.60\\
								$5$  & AIM \cite{vu2018perception}			& 3.321	& 11.37	&& $5$ 	& BOE \cite{Navarrete2018multi}			& 2.484 & 12.50	&& $5$  & TTI (Ours)		& 2.040	& 13.17\\
								$6$  & DSP-whu		& 3.728	& 11.45	&& $6$ 	& AIM \cite{vu2018perception}			& 2.600 & 12.42	&& $6$  & Haiyun \cite{luo2018bi} & 2.077	& 15.95\\
								$7*$ & BOE \cite{Navarrete2018multi} & 3.817	& 11.50	&& $7$ 	& REC-SR \cite{kuldeep2018scale} & 2.635 & 12.37	&& $7$  & gayNet		& 2.104	& 15.88\\
								$7*$ & REC-SR \cite{kuldeep2018scale}		& 3.831	& 11.46	&& $8$ 	& DSP-whu		& 2.660 & 12.24 && $8$  & DSP-whu		& 2.114	& 15.93\\
								$9$  & Haiyun \cite{luo2018bi}	& 4.440	& 11.19	&& $9$ 	& XYN			& 2.946 & 12.23	&& $9$  & MCML	& 2.136	& 13.44\\
		\bottomrule
	\end{tabu}
	\label{tab:results_perceptual}
\end{table*}

The generator loss in this experiment is composed of four loss terms, following~\cite{sajjadi2016enhancenet}: MSE, VGG, Style, and Adversarial loss. 
\begin{equation}
L_{G} = w_1*L_{mse} + w_2*L_{vgg} + w_3*L_{adv} + w_4*L_{style} 
\end{equation}
\begin{itemize}
\item MSE loss is pixel-wise loss which calculated in the image space $L_{mse} = || I^h - I^{sr} ||^2_2$. 
\item VGG loss is calculated in the feature space using pretrained VGG19 network~\cite{simonyan2014very} on multiple layers. This loss was originally proposed by~\cite{johnson2016perceptual, dosovitskiy2016generating}. Both $I^h$ and $I^{sr}$ are first mapped into a feature space by differentiable functions $f_{i}$ from VGG multiple max-pool layers ${(i = {2, 3, 4, 5})}$ then sum up each layer distances. $L_{vgg} = \sum\limits_{i={2}}^5 || f_i(I^h) - f_i(I^{sr}) ||^2_2$.
\item Adversarial loss. $L_{adv} = -\texttt{log} (D(G(I^{l})))$, where
  $D(x)$ is the probability assigned by $D$ to $x$ being a real HR image.
\item Style loss is used to generate high quality textures. This loss was originally proposed by~\cite{gatys2016image} which is later modified by~\cite{sajjadi2016enhancenet}. Style loss uses the same differentiable function $f$ as in VGG loss. $L_{style} = \sum\limits_{i={2}}^5 || \theta(f_{i}(I^h)) - \theta(f_{i}(I^{sr})) ||^2_2$ where Gram matrix $\theta(F)=FF^T \in \mathbb{R}^{n\times n}$.
\end{itemize}

The training objective for $D$ is 
\[
  L_D = -\texttt{log} (D(I^{h})) - \texttt{log}(1-D(G(I^{l}))).
\]
As is common in training adversarial networks, we alternate between
stages of training $G$ and training $D$.
We use pre-trained DBPN model which optimized by MSE loss only, then fine-tuned with the perceptual loss. We use batch size of 4 with size $60 \times 60$ for LR image, while HR image size is $240 \times 240$. The learning rate is initialized to $1e-4$ for all layers for $2\times10^5$ iteration using Adam with momentum to $0.9$. 

This method was included in the challenge associated with
PIRM2018~\cite{pirm2018}, in conjunction with ECCV 2018.
In the challenge, evaluation was conducted in three disjoint
\emph{regimes} defined by thresholds on the RMSE; the intuition behind
this is the natural tradeoff between RMSE and perceptual quality of
the reconstruction. The latter is measured by combining the quality measures of Ma~\cite{ma2017learning} and NIQE~\cite{mittal2013making} as below,
\begin{equation}
\texttt{Perceptual index} = 1/2((10-Ma)+NIQE).
\end{equation}

The three regimes correspond to Region 1: RMSE $\le 11.5$ , Region 2:
$11.5 <$ RMSE $\le 12.5$, and Region 3: $12.5 <$ RMSE $\le 16$. We
select optimal parameter settings for each regime. 
This process yields
\begin{itemize}
\item Region 1 ($w_1: 0.5, w_2: 0.05, w_3: 0.001, w_4: 1$)
\item Region 2 ($w_1: 0.1, w_2: 0.2, w_3: 0.001, w_4: 1$)
\item Region 3 ($w_1: 0.03, w_2: 0.2, w_3: 0.001, w_4: 10$)
\end{itemize}

Our method achieved $1^{st}$ place on Region 2, $3^{rd}$ place on
Region 1, and $5^{th}$ place on Region 3~\cite{pirm2018} as shown in Table~\ref{tab:results_perceptual}. In Region 3, it shows very competitive results where we got $5^{th}$, however, it is noted that our method has the lowest RMSE among other top 5 performers which means the image has less distortion or hallucination w.r.t the original image.

\begin{figure*}[t!]
\centering
\includegraphics[width=16cm]{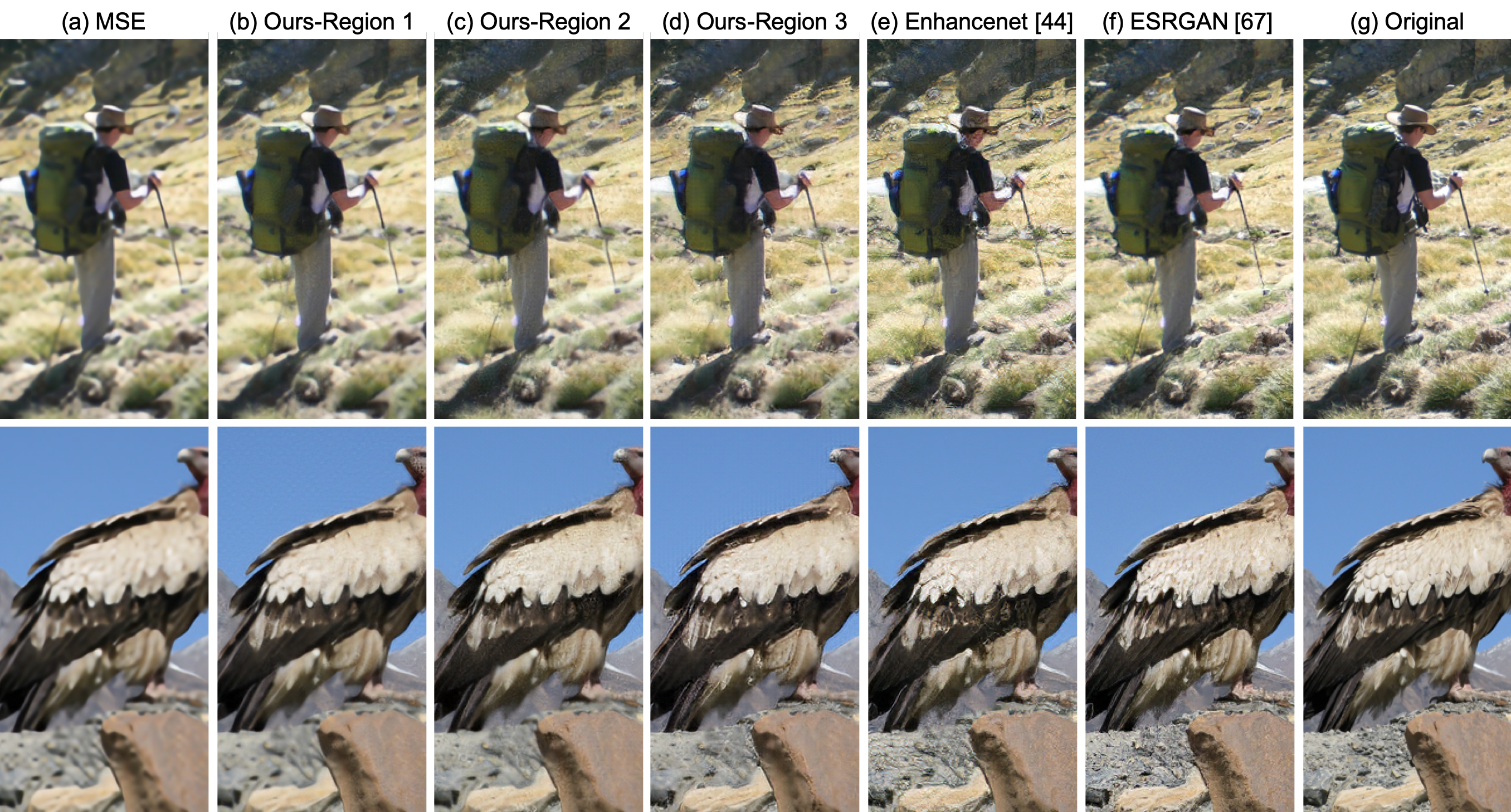}
\caption{Results of DBPN with perceptual loss compare with other methods.}
\label{figure:perceptual_img}
\end{figure*}

We show qualitative results from our method which is shown in Fig.~\ref{figure:perceptual_img}. 
It can be seen that there are significant improvement on high quality texture on each region compare to MSE-optimized SR image.
ESRGAN~\cite{wang2018esrgan}, the winner of PIRM2019 on Region 3, gets the best perceptual results among other methods.
However, our proposal contains less noise among other methods (the smallest RMSE) while maintaining good perceptual quality on Region 3.
\section{Conclusion}
We have proposed Deep Back-Projection Networks for Single Image Super-resolution which is the winner of two single image SR challenge (NTIRE2018 and PIRM2018). Unlike the previous methods which predict the SR image in a feed-forward manner, our proposed networks focus to directly increase the SR features using multiple up- and down-sampling stages and feed the error predictions on each depth in the networks to revise the sampling results, then, accumulates the self-correcting features from each upsampling stage to create SR image. We use error feedbacks from the up- and down-scaling steps to guide the network to achieve a better result. The results show the effectiveness of the proposed network compares to other state-of-the-art methods. Moreover, our proposed network successfully outperforms other state-of-the-art methods on large scaling factors such as $8\times$ enlargement. We also show that DBPN can be modified into several variants to follow the latest deep learning trends to improve its performance.
\ifCLASSOPTIONcompsoc
  \section*{Acknowledgments}
\else
  \section*{Acknowledgment}
\fi
This work was partly supported by JSPS KAKENHI Grant Number 19K12129
and by AFOSR Center of Excellence in Efficient and Robust Machine
Learning, Award FA9550-18-1-0166.

\ifCLASSOPTIONcaptionsoff
  \newpage
\fi



%
{\small
\bibliographystyle{IEEEtran}
\bibliography{egbib}
}

%

\begin{IEEEbiography}[{\includegraphics[width=1in,height=1.25in,clip,keepaspectratio]{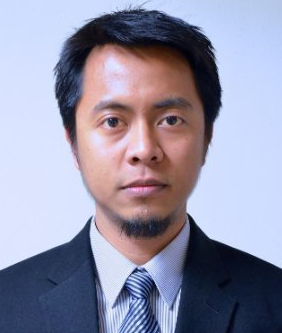}}]{Muhammad Haris}
Muhammad Haris received S. Kom (Bachelor of Computer Science) from the Faculty of Computer Science, University of Indonesia, Depok, Indonesia, in 2009. Then, he received the M. Eng and Dr. Eng degree from Department of Intelligent Interaction Technologies, University of Tsukuba, Japan, in 2014 and 2017, respectively, under the supervision of Dr. Hajime Nobuhara. He worked as postdoctoral fellow in Intelligent Information Media Laboratory, Toyota Technological Institute with Prof. Norimichi Ukita. Currently, he is working as AI Research Manager at BukaLapak.
\end{IEEEbiography}

\begin{IEEEbiography}[{\includegraphics[width=1in,height=1.25in,clip,keepaspectratio]{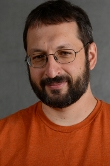}}]{Greg Shakhnarovich}
Greg Shakhnarovich has been faculty member at TTI-Chicago since
2008. He received his BSc degree in Computer Science and Mathematics from
the Hebrew University in Jerusalem, Israel, in 1994, and a MSc degree
in Computer Science from the Technion, Israel, in 2000. Prior to
joining TTIC Greg was a Postdoctoral Research Associate at Brown
University, collaborating with researchers at the Computer Science
Department and the Brain Sciences program there. Greg's research
interests lie broadly in computer vision and machine learning.
\end{IEEEbiography}

\begin{IEEEbiography}[{\includegraphics[width=1in,height=1.25in,clip,keepaspectratio]{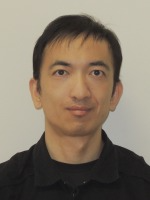}}]{Norimichi Ukita}
Norimichi Ukita is a professor at
the graduate school of engineering, Toyota Technological
Institute, Japan (TTI-J). He received
the B.E. and M.E. degrees in information engineering
from Okayama University, Japan, in
1996 and 1998, respectively, and the Ph.D degree
in Informatics from Kyoto University, Japan,
in 2001. After working for five years as an
assistant professor at NAIST, he became an
associate professor in 2007 and moved to TTIJ
in 2016. He was a research scientist of Precursory Research for
Embryonic Science and Technology, Japan Science and Technology
Agency (JST), during 2002 - 2006. He was a visiting research scientist
at Carnegie Mellon University during
2007-2009. He currently works also
at the Cybermedia center of Osaka University as a guest professor.
His main research interests are object detection/tracking and human
pose/shape estimation. He is a member of the IEEE.
\end{IEEEbiography}
\vfill





\end{document}